%% file: main_rlc.tex
\documentclass[10pt]{article} 

\usepackage[accepted]{rlj}           

\usepackage{amssymb}            
\usepackage{mathtools}          
\usepackage{mathrsfs}           
\usepackage{graphicx}           
\usepackage{subcaption}         
\usepackage[space]{grffile}     
\usepackage{url}                
\usepackage{lipsum}             

\hypersetup{
    colorlinks=true,
    linkcolor=black,
    citecolor=black,
    filecolor=magenta,
    urlcolor=darkgray,
}

\usepackage{wrapfig}
\usepackage{fancyhdr}
\usepackage{graphicx}
\usepackage{amsmath,amsfonts,amssymb,amsthm}
\usepackage{float}
\usepackage[mathscr]{euscript}
\usepackage[dvipsnames]{xcolor}
\usepackage{multicol}
\usepackage[ruled]{algorithm}
\usepackage[noend]{algorithmic}
\usepackage{stmaryrd}
\usepackage{thm-restate} 
\usepackage[nameinlink,capitalize]{cleveref}

\newif\ifshowcomments
\showcommentstrue   
\usepackage[normalem]{ulem}

\usepackage{tikz}

\ifshowcomments
  \newcommand{\sorina}[1]{\textcolor{red}{Sorina: #1}}
  \newcommand{\john}[1]{\textcolor{blue}{John: #1}}
  \newcommand{\martha}[1]{\textcolor{magenta}{Martha: #1}}
  \newcommand{\joseph}[1]{\textcolor{purple}{Joseph: #1}}
  \newcommand{\patrick}[1]{\textcolor{teal}{Patrick: #1}}
  \newcommand{\jun}[1]{\textcolor{Maroon}{Jun: #1}}
  \newcommand{\kris}[1]{\textcolor{orange}{Kris: #1}}
  \newcommand{\mohamed}[1]{\textcolor{teal}{Mohamed: #1}}
  \newcommand{\khurram}[1]{\textcolor{pink}{Khurram: #1}}
  
\else
  \newcommand{\sorina}[1]{}
  \newcommand{\john}[1]{}
  \newcommand{\martha}[1]{}
  \newcommand{\joseph}[1]{}
  \newcommand{\patrick}[1]{}
  \newcommand{\jun}[1]{}
  \newcommand{\kris}[1]{}
  \newcommand{\mohamed}[1]{}
  \newcommand{\khurram}[1]{}
\fi
\usepackage{makecell}

  \newcommand{\revision}[1]{\textcolor{black}{#1}}

\usepackage{pdfpages}

\title{The Open Ant: A Robot Platform for Reinforcement Learning Research}

\setrunningtitle{The Open Ant: A Robot Platform for Reinforcement Learning Research}

\author{
Elena Sorina Lupu\textsuperscript{1*}, 
Patrick Spieler\textsuperscript{*}, 
Khurram Javed\textsuperscript{2}, 
Kris De Asis\textsuperscript{1}, 
John D. Martin\textsuperscript{1, 3}, 
Martha Steenstrup\textsuperscript{3}, 
Joseph Modayil\textsuperscript{1, 2} 
}

\emails{}

\affiliations{
$^{1}$\textbf{Openmind Research Institute, Canada}\\
$^{2}$\textbf{Keen Technologies}\\
$^{3}$\textbf{Department of Computing Science, University of Alberta, Canada}\\

\par 
$^*$equal contribution
}

\contribution{
We developed a physical robot platform with an accompanying simulation, designed to support reinforcement learning researchers who are not accustomed to working with robots.
The robot is inspired by the Gymnasium Ant used in continuous-control research, but several alterations were required to make a useful physical research platform.
The hardware design and software are released to the community as an open-source platform.}
{
The Gymnasium Ant~\citep{Schulman2015HighDimensionalCC} is a popular simulated domain, with an abstract physical design. Reinforcement learning algorithm research is commonly conducted with simulated domains, though the community has used several robot platforms to validate their algorithms on embodied systems. The proposed platform is intended to support such efforts.
}

\contribution{
We demonstrate that this platform is compatible with multiple reinforcement learning methods.  The platform supports both simple and \revision{complicated} RL algorithms.  The platform supports learning directly from the hardware experience and also policy transfer from the simulator to reality.
}
{
The reinforcement learning community has adopted a wide range of methods and research directions. By successfully demonstrating the use of different RL algorithms and methods with this platform, we make it easier for RL researchers to adopt the platform.  
}

\contribution{
We show that the platform supports a nimble ecosystem for RL experimentation, with the rapid on-boarding of inexperienced users and rapid revisions when hardware failures are encountered.  The platform was used for RL experimentation by multiple researchers with limited prior experience in robotics. The platform supports rapid revision to encountered problems, through the use of 3D printing and commercially available components.
}
{
RL researchers who used robots previously experienced challenges with both the initial adoption of a new platform, and with maintaining a robot as a research platform.  
}


\summary{
Reinforcement learning (RL) research has demonstrated success in both physical and simulated domains; however, the predominant methodology remains rooted in simulations. The predominance of simulations makes translating research to physical reality uncertain for both algorithms and researchers.
We propose a physical platform that is designed to simplify the transition. 
In this paper, we present the Open Ant: a physical variant of the commonly used Gymnasium Ant environment, along with a simulation. 
We demonstrate that competent walking policies can be learned from scratch in approximately one hour directly from the physical robot's experience for two substantially different RL algorithms: SARSA($\lambda$) and Soft Actor-Critic (SAC). 
Separately, we show policies that were learned in simulation transfer to reality. 
We also examine how well the platform supports a nimble experimental ecosystem. Specifically, we observe the speed with which new users from diverse backgrounds achieve their first success with the platform, and how easily the platform can be repaired and updated when hardware issues arise.
Both the hardware design and software are available as open-source \revision{on \href{https://github.com/Openmind-Research-Institute/open-ant}{GitHub}} for ease of customization.
In summary, we advocate for the use of the Open Ant for RL researchers who frequently use simulated environments, so they can more easily include robot experiments in their evaluations.
}

\begin{document}

\makeCover  
\maketitle  

\begin{abstract}
Reinforcement learning (RL) research has demonstrated success in both physical and simulated domains; however, the predominant methodology remains rooted in simulations. The predominance of simulations makes translating research to physical reality uncertain for both algorithms and researchers.
We propose a physical platform that is designed to simplify the transition. 
In this paper, we present the Open Ant: a physical variant of the commonly used Gymnasium Ant environment, along with a simulation. 
We demonstrate that competent walking policies can be learned from scratch in approximately one hour directly from the physical robot's experience for two substantially different RL algorithms: SARSA($\lambda$) and Soft Actor-Critic (SAC). 
Separately, we show policies that were learned in simulation transfer to reality. 
We also examine how well the platform supports a nimble experimental ecosystem. Specifically, we observe the speed with which new users from diverse backgrounds achieve their first success with the platform, and how easily the platform can be repaired and updated when hardware issues arise.
Both the hardware design and software are available as open-source \revision{on \href{https://github.com/Openmind-Research-Institute/open-ant}{GitHub}} for ease of customization.
In summary, we advocate for the use of the Open Ant for RL researchers who frequently use simulated environments, so they can more easily include robot experiments in their evaluations.
\end{abstract}

\section{Introduction}

Reinforcement learning (RL) has been successful in both physical and simulated domains, though many successes rely heavily on accurate simulations. 
Several of the notable successes include games such as Backgammon~\citep{tesauro1995temporal} and Go~\citep{silver2016mastering}, where accurate and computationally efficient simulators can capture the exact dynamics. Success in complex online games can introduce differences between training and deployment environments~\citep{wurman2022outracing}.  Successes in complex physical domains has often relied on human expertise to select the most relevant dynamics to simulate, with policies trained in simulation prior to a deployment phase, as seen in balloon navigation~\citep{bellemare2020autonomous}, fusion plasma control~\citep{degrave2022magnetic}, \revision{or gravitational wave detectors~\citep{doi:10.1126/science.adw1291}}.

Reinforcement learning research might be a victim of its own success with simulations, as relatively few studies focus on learning directly in physical reality. The ability to learn directly from physical experience has ample evidence in animal learning experiments, which served as motivation for early computational RL algorithms~\citep{barto1983neuronlike}\revision{, with a modified version demonstrated on hardware~\citep{287219}}. This early motivation was followed by evidence that RL algorithms provide computational models for the activity of dopamine neurons~\citep{montague1996framework, schultz1997neural}.
In addition to animal studies, many engineered systems have demonstrated the use of RL algorithms to learn behaviors directly from physical experience without the need for a simulator, with early examples on locomotion~\citep{robot_dog_peter_stone,tedrake2004stochastic} and later work demonstrating the use of a single RL algorithm across multiple robots and problem formulations~\citep{DBLP:journals/corr/abs-1809-07731, wu2023daydreamer}.  

Despite repeated success with RL algorithms on physical robots, researchers with primarily simulation-based experience encounter significant challenges when experimenting with physical robots. 
One challenge is that the overhead (e.g., cost and time) for experiments with physical robotics is substantially higher than using simulations alone. 

Even in situations where adequate lab resources and expertise are available, the long delay caused by the assembly of the robot and troubleshooting before the first successful experiment can exhaust a considerable portion of a researcher's typical residency. 
Consequently, experts may graduate and depart research labs shortly after successful experiments, resulting in slower experiment iteration and difficulties with future re-use.

\begin{figure}[t]
    \centering
    \includegraphics[width=\linewidth]{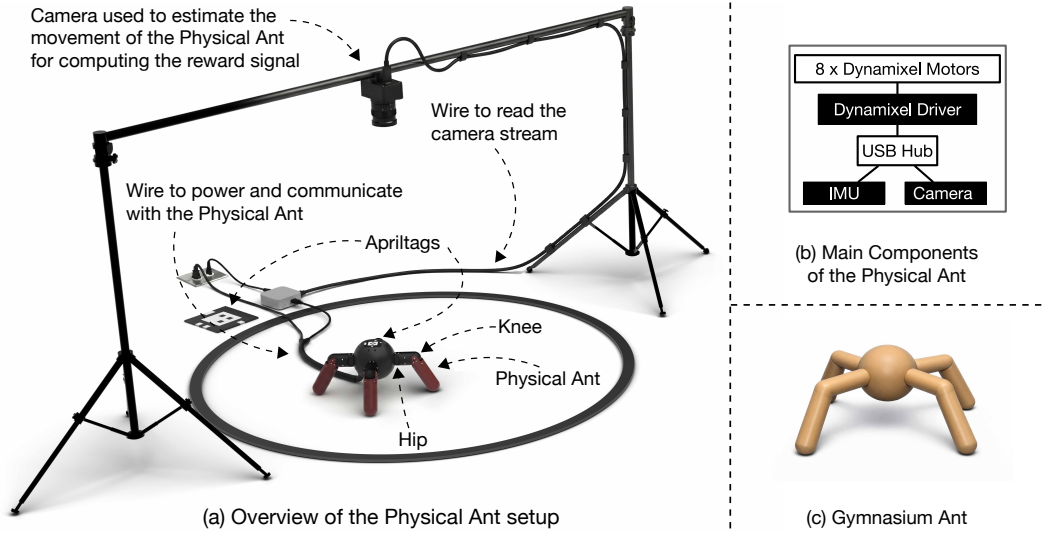}
    \caption{\textbf{Physical Ant platform, learning arena, and system overview.} (a)  
    An overhead webcam tracks the fiducial markers to compute reward signals and the heading vector of the ant. The robot is connected by cables to AC power and to an external computer where the agent is running. 
    (b) The main components of the Physical Ant. (c) The Gymnasium Ant~\citep{Schulman2015HighDimensionalCC,towers2025gymnasiumstandardinterfacereinforcement}, which was the inspiration for the Physical Ant.}
\label{fig:system_architecture}
\end{figure}

\paragraph{Contributions.}

We present the Open Ant, an open-source research robot platform available on \revision{ \href{https://github.com/Openmind-Research-Institute/open-ant}{GitHub}}. 
The robot comes with a MuJoCo simulation, which can be used to learn competent policies.
The robot body is designed to look like the widely used Gymnasium Ant environment~\citep{towers2025gymnasiumstandardinterfacereinforcement}. 
The Open Ant makes physical robot experiments an easier addition to a standard RL research pipeline. 
The robot is designed to be built and maintained by AI researchers without backgrounds in robotics engineering. 
The robot is designed to withstand the substantial wear incurred during RL exploration, and to be easy to repair when components are damaged. 

The learning experiments presented here are designed to yield positive results within an hour; furthermore, we demonstrate that the robot platform provides reliable, repeatable results across different RL algorithms. 
In these ways, we envision this platform lowering the barrier to physical robotics research for the broader RL community.

\section{Platform Motivation and Background}\label{sec:motivation_and_background}

Our overall goal is to create a research platform that \revision{simplifies the process for researchers to extend their simulation results in RL research.}
Based on our team's experience on previous robot platforms, we identified several core criteria: the robot should closely mirror an existing simulated domain (see \Cref{fig:system_architecture}), remain affordable, and integrate with standard RL software tools (see~\Cref{sec:results}). Furthermore, running experiments should require minimal intervention, support diverse RL methods, and yield results on hardware within timescales comparable to simulation (see~\Cref{sec:results}).
Finally the overall platform should be accessible to newcomers and relatively easy to modify and maintain (\Cref{sec:suitability_for_RL}).
Although many prior works have demonstrated some of these capabilities, we are not aware of a platform that provides all of them.
As such, our goal is not to produce a static benchmark, but to introduce a platform that the research community can flexibly revise over time---similar to the Arcade Learning Environment~\citep{bellemare2013arcade}.

In addition to making robot experiments easier, we seek a platform that supports RL researchers to better study fundamental questions that arise when algorithms learn from physical experience. Invited talks at the 2025 RL Conference by \cite{kaelbling2025rational} and \cite{sutton2025oak}  highlighted the complementary considerations for RL algorithms that happen at \textit{design-time} (before regular deployment) and \textit{run-time} (during regular deployment). 
Design-time knowledge of the problem structure already informs many facets of a robot system using RL, including the selection of the observation and action spaces~\citep{mahmood2018settingreinforcementlearningtask}, the creation of the initial behavior policies \citep{DBLP:journals/corr/abs-1812-06298}, the physical design of the robot's body, and the selection of appropriate simulators. 
In contrast, run-time knowledge can only be acquired in deployment. During run-time, the agent may encounter phenomena that are either unknown or imperfectly modeled at design-time.

\revision{Traditional adaptive control~\citep{slotine1991applied} addresses run-time learning by continuously updating a linear controller to compensate for changes in the environment parameters while the robot remains in operation. Examples of online learning using adaptive control on hardware include remote-controlled airplanes~\citep{Shi2020Adaptive}, quadcopters~\citep{O_Connell_2022}, and ground vehicles~\citep{10.1109/TRO.2024.3475212}. Reinforcement learning shares this same online learning/adaptation property but extends it beyond tracking or stabilization, enabling complex decision-making through interaction with the environment~\citep{sutton2018reinforcement}. Physical robots provide a natural setting for studying algorithms that learn continually during deployment, where adaptation must occur from ongoing experience rather than repeated offline retraining.}

In practice, most RL experiments are conducted in a lab without deployment \revision{(sim-to-real)}; however, we aim to study algorithms that can adapt behavior in deployment, outside of simulators and with limited researcher intervention. 
A primary constraint of the run-time setting is learning while behaving, without pausing the environment. 
Moreover, the robot's physical experience differs from a simulation: a ground-truth simulator state is unavailable, and physical effects like overheating are difficult to model. 
Additional considerations include ensuring learning algorithms are compatible with local compute, communication, and timing constraints, as well as limiting the need for manual episodic resets. Our aim is to make these run-time conditions accessible and practical for RL researchers through a single, integrated robotic platform for learning from physical experience.

Although several papers have demonstrated run-time reinforcement learning on physical robots, it remains unclear whether these platforms are easily adopted by current RL researchers.  One fruitful approach is to design custom robots that mirror existing simulated domains. 
For instance, NoodleBot~\citep{berrueta2024maximum} replicates the Gymnasium Swimmer benchmark, while RealAnt~\citep{DBLP:journals/corr/abs-2011-03085} is a quadrupedal platform inspired by the Gymnasium Ant.
\revision{In contrast to ours, the RealAnt requires soldering and overhead camera calibration, and lacks clear guidance on setup time and reproducibility. In addition, its morphology also differs from the standard MuJoCo Ant model.}
Run-time reinforcement learning has also been demonstrated on commercial platforms. 
For example,~\cite{smith2022walkparklearningwalk} show run-time policy learning on the Unitree A1 quadruped, and \cite{robot_dog_peter_stone} demonstrate a policy gradient RL algorithm that learns the parameters of a walking gait for the Sony Aibo quadruped. 
More recently,~\cite{preiss2025fastnonepisodicadaptivetuning} demonstrate non-episodic, online policy optimization on a Crazyflie open-source quadcopter~\citep{7989376}.
Even more closely aligned with our objectives,
\cite{mahmood2018settingreinforcementlearningtask}~previously demonstrated the
 instantiation of a Gymnasium Reacher environment with a commercial UR5 robot. We chose to pursue a different platform from the Reacher to access more domain complexity (higher dimensional observations and actions, mobility with contact dynamics) at a fraction of the cost.
Commercial platforms can offer a relatively low barrier of entry for RL researchers familiar with simulations; however, they carry risks of product obsolescence, high purchase costs, long and costly repairs, and unfamiliar or rigid software interfaces which may hinder certain lines of research.

\section{The Open Ant Platform}
In this section, we introduce the Open Ant (\Cref{fig:system_architecture}): an \textit{open-source} quadruped platform for RL and robotics research.
Our platform includes both physical and simulated robot bodies inspired by the Gymnasium Ant \citep{Schulman2015HighDimensionalCC}, which serves as a popular benchmark for continuous-control research \citep{towers2025gymnasiumstandardinterfacereinforcement}. Our physical robot body is called Physical Ant and the simulation is called Simulated Ant. Our platform is designed for researchers to easily move between simulation and reality when experimenting.


\subsection{Design Overview}
\paragraph{Design Principles.} Our design adheres to three fabrication principles.
First, the design \textit{prioritizes simplicity and ease of assembly} by avoiding any need for soldering and relying exclusively on commercial off-the-shelf (COTS) components that are widely available. Second, the robot must \textit{support long-term learning} without interruption; thus it is powered directly from an AC wall outlet, bypassing the maintenance overhead of rechargeable batteries.
Third, the robot requires \textit{no special equipment}; it directly connects to a personal computer via USB, eliminating the need for specialized embedded systems code or complex network configurations.

\begin{table}[t]
    \caption{Comparison of the Gymnasium Ant and the Physical Ant platform.}
    \centering
    \begin{tabular}{l|c|c|c|c|c|c}
        Ant Type  & \makecell{Torso \\ Diameter \\ $[$m$]$ } & \makecell{Upper \\ Leg \\ $[$m$]$ }& \makecell{Lower \\ Leg \\ $[$m$]$ } & \makecell{Total \\ Mass \\ $[$kg$]$} & \makecell{Hip Motor} & \makecell{Knee Motor}  \\
        \hline 
        Physical Ant &  0.15 & 0.08 & 0.16 & 1.20 & XC430-W240 & XM430-W210  \\
        Gymnasium Ant  & 0.50 & 0.28 & 0.56 & 0.91 & N/A & N/A \\
    \end{tabular}
    \label{tab:ant_comparison}
\end{table}

\paragraph{Physical Attributes.}~\label{sec:geom}
The Physical Ant is one third the scale of the Gymnasium Ant \citep{Schulman2015HighDimensionalCC}. By default, the Gymnasium Ant stands approximately 75 cm tall with a total mass of 0.9 kg, its torso is 0.50 m in diameter; the upper leg segment measures 0.28 m and the lower leg segment 0.56 m. These dimensions imply an extremely low mass-to-size ratio that would be difficult to realize physically with conventional materials.
The Physical Ant's reduced size results in a relatively higher mass-to-length ratio, improving both its robustness and ease of manufacturing. See \autoref{tab:ant_comparison} for \revision{the} design details.

\paragraph{Construction.}
The robot consists of a spherical body (torso) housing the onboard electronics and four articulated legs.  The torso and legs are 3D printed, enabling rapid revision and localized repairs.
Each leg is equipped with two metal-geared Dynamixel actuators: one at the hip (XC430-W240) and one at the knee (XM430-W210). 
The spherical body shell houses the onboard electronics, including an IMU, a \revision{USB communication converter}, and a USB hub that interfaces these components with an external computer (\Cref{fig:system_architecture}, (b)). 
Additionally, the Physical Ant features an on-board camera, to support future vision-based learning tasks. \revision{The complete list of components is available on our \href{https://github.com/Openmind-Research-Institute/open-ant}{GitHub} repository.}
The robot communicates via USB with an external computing platform, a design choice that allows users to flexibly select their preferred compute hardware.
\revision{As a result, researchers can develop and run their algorithms without needing to adapt them to the computational constraints of an onboard platform.}
Power is provided via a wall-plug power supply, and the feet are equipped with 3D-printed thermoplastic polyurethane (TPU) ``socks''  to increase ground friction. The total cost for all the components is approximately USD 2200, excluding 3D-printing filament.


There exists a possibility to replace the more expensive motors (XM430-W210) with plastic-geared motors (XL430-W250) on both joints, which we call the Physical Ant Lite. For this lower-cost variant, the overall price is approximately USD 500 (excluding 3D printed parts).
While more affordable, these actuators are more susceptible to wear and overheating.
We therefore recommend that interested users carefully review the thermal analysis (\Cref{sec:thermal_analysis}), as well as the discussed electrical and mechanical improvements (\Cref{sec:electrical_analysis} and \Cref{sec:suitability_for_RL}) \revision{before deciding on the best motors for their needs.}

\subsection{Differences and Similarities between the Gymnasium and the Open Ant Environments}
We highlight several aspects of the Gymnasium Ant environment \citep{Schulman2015HighDimensionalCC} that motivated our design. 
The Physical Ant adopts the same software interface as the Gymnasium Ant, while varying in the specific choices for its sensing, actuation, and reward.

\paragraph{Observations.}\label{sec:obs} 
Compared to its Gymnasium counterpart, the Physical Ant has a compact observation space. Its observations have twenty-four dimensions, consisting of joint angles (8), joint angular velocities (8), the torso’s angular velocity (3), and linear acceleration (3) measured by an onboard IMU. In addition, \revision{we augment the observation with a} two-dimensional unit vector encoding the angle between the heading vector and the goal direction (\autoref{eq:switching}) \revision{for the} back-and-forth task \revision{presented below}. Lastly, the platform provides interfaces for augmenting the observation space with other sensor modalities, such as motor loads and motor temperatures.
In contrast, the observation space of Gymnasium Ant is 105 dimensions, composed of body-part positions, velocities, and external forces acting on each segment. Observations exclude inertial planar position. 
Many of these observations in the Gymnasium Ant, such as the external forces on the body parts, are difficult to obtain on hardware.

\paragraph{Actions.}\label{sec:actions}
The Physical Ant's action space has eight continuously-valued, bounded \revision{dimensions}. Actions represent position commands for the hip and knee joints of all four legs. Each command specifies a desired joint position to a Dynamixel actuator. For the Simulated Ant \revision{in MuJoCo}, desired positions are tracked using proportional–derivative control whose gains and force limits were chosen to match the physical hardware \revision{(see~\Cref{sec:sys_id} for system identification)}.
In comparison, the Gymnasium Ant adopts a torque control strategy.
\revision{Some prior real-world RL work uses position control~\citep{doi:10.1126/scirobotics.abk2822}. The software control interface can be changed to use position, velocity, or torque commands for other experiments.} 

\begin{figure}[t]
    \centering
    \includegraphics[width=\linewidth]{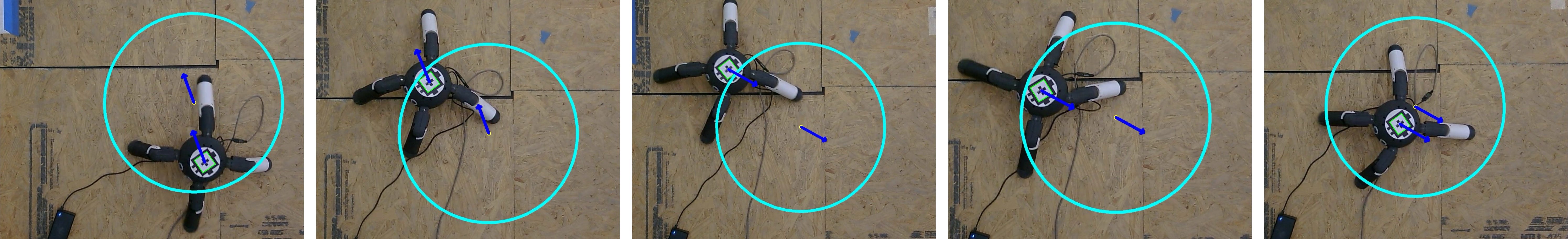}
    \caption{\textbf{Overhead snapshots during run-time learning from physical experience}. The {\color{cyan}cyan} circle represents the boundary where the reward direction (the {\color{blue}blue} arrow) changes direction if the Physical Ant exceeds it. In the first figure, the robot travels towards 11 o'clock. In the second figure, it approaches the boundary. In the third figure, the reward direction flips and in the last two figures, we see the ant traveling in the direction of the reward direction.}
    \label{fig:hardware_back_and_forth}
\end{figure}

\paragraph{Rewards and Tasks.}\label{sec:rewards}
The Gymnasium Ant’s default task is to move forward as quickly as possible.
The reward function is a sum of multiple terms: a reward for staying upright, \revision{called a \textit{healthy reward},} a \textit{forward reward} for \revision{making} forward progress, a \textit{control cost} penalizing large actions, and an optional \textit{contact cost} penalizing large external forces. The Gymnasium Ant also has episodic resets back to a starting configuration, which truncates episodes after long trajectories or on arrival to an unhealthy state.  
In contrast, we aim for learning in a non-episodic setting on the Physical Ant with a \textit{simple reward formulation} based on progress alone. 

Towards this end, we design a non-episodic task of moving back-and-forth. 
We define the robot's planar position at time $t\in\mathbb{N}$ to be $\mathbf{p}_t=[
x_t, y_t ] \in \mathbb{R}^2,$ measured in the inertial (world) frame and a reward direction unit vector $\mathbf{u}_t\in\mathbb{R}^2$ \revision{defined in Equation~\ref{eq:switching}}.
The instantaneous reward $r_t\in\mathbb{R}$ is given by the projection of the position displacement onto the current reward direction $\mathbf{u}_t$
\begin{equation}\label{eq:back_and_forth_reward}
r_t = (\mathbf{p}_t - \mathbf{p}_{t-1})^\top \mathbf{u}_t.
\end{equation}
We further define a circle of radius $R>0$ centered at $\mathbf{o}\in\mathbb{R}^2$. 
\revision{When the robot's position is outside the circle and it has made it to the half plane opposite from the previous reward direction update, a change in reward direction is triggered that compels the robot `bounce back' from the circle, as follows:}
\begin{equation}\label{eq:switching}
\mathbf{u}_{t+1} =
\begin{cases}
\dfrac{\mathbf{o} - \mathbf{p}_t}{\|\mathbf{o} - \mathbf{p}_t\|_2},
& \text{if } 
\|\mathbf{p}_t - \mathbf{o}\|_2 > R
\ \text{and}\ 
(\mathbf{p}_t - \mathbf{o})^\top \mathbf{u}_{t} > 0, \\[6pt]

\mathbf{u}_{t},
& \text{otherwise}.
\end{cases}
\end{equation}
An illustration of the switching behavior in~\autoref{eq:switching} is shown in~\Cref{fig:hardware_back_and_forth}. 
The advantage of the back-and-forth task, in contrast to the forward task specified in the Gymnasium Ant, is that it does not require the user to bring a physical robot back to the origin when it reaches the edge of the learning arena. 
In this way, the robot learns forever, with minimal user intervention.

\paragraph{Learning Arena.}\label{sec:learning_arena}
The robot pursues the back-and-forth walking task in an arena similar to the one rendered in~\Cref{fig:system_architecture}(a). 
An overhead webcam supported by a tripod system tracks two fiducial markers~\citep{olson-april-11}: one placed on the ant's torso and another one placed on the ground. 
The system provides the position $\mathbf{p}$ used to compute the reward in \autoref{eq:back_and_forth_reward} and the heading.


\subsection{Measuring Performance.}\label{sec:measuring_performance}
The performance of the learned policy is measured with the \textit{average reward per second} defined as
\begin{equation}\label{eq:average_reward}
\bar{r}_t=\frac{1}{N} \sum_{k=t-N+1}^t \frac{r_k}{\Delta t},
\end{equation}
where $r_k\in\mathbb{R}$ is the instantaneous reward, $N\in\mathbb{N}$ is the number of steps contained in the averaging window (corresponding to a user-defined window length), $\Delta t\in\mathbb{R}_{>0}$ is the time duration of one environment step, and $t\in\mathbb{N}$ is the current time step.

\section{Reinforcement Learning Methods}\label{sec:benchmarking_RL}

We provide demonstrations of learning from physical experience for two representative RL algorithms, under the back-and-forth walking task described in the previous section. 
The first algorithm is an on-policy action-value method: SARSA$(\lambda)$ with linear approximation of the value function~\citep{sutton2018reinforcement}. SARSA$(\lambda)$ has a small number of parameters and is easy to implement.
The second algorithm is Soft Actor-Critic (SAC)~\citep{sac}, a deep RL off-policy entropy-regularized method for continuous control problems. 
For each algorithm, we describe the implementation details in this section and report the corresponding experimental results in~\Cref{sec:results}.
These two demonstrations show that learning from physical experience is feasible. They also provide insight into the behavior and limitations of run-time learning with the Physical Ant platform.

\subsection{Discrete Action RL: SARSA($\lambda$)} \label{paragraph:sarsa}

SARSA($\lambda$) is an on-policy RL algorithm where an agent interacts with the environment, learning an action-value function that defines the policy (for example, using $\varepsilon$-greedy \revision{action selection}). The algorithm is defined with a discrete set of actions, so a transformation is needed from the robot's continuous action space.

Before applying SARSA($\lambda$) as a learning algorithm for the Open Ant, we first defined a slower and more abstract agent-environment interface.  The abstract learning agent interface operated at a slower rate (0.5 s) than what was used for communication to the robot (0.05 s), where each agent timestep lasted for $\tau=10$ robot interactions. \revision{These timing values were a design choice.}
On each agent timestep, the agent received the latest observation from the robot. The intermediate rewards from the robot were accumulated before being sent to the learning agent, where the agent's reward between the robot timesteps $t$ and $t+\tau$ is the discounted sum
$r_{t+1} + \gamma r_{t+2} + \ldots + \gamma^{\tau-1} r_{t+\tau}$, where $\gamma$ is the discount factor. 
For simplicity, we use the discounted formulation of SARSA($\lambda$) rather than an average-reward formulation. The robot's continuous command space was abstracted into a discrete set of long duration motion primitives, which formed the learning agent's actions.
We defined each agent action $\bar{\mathbf{a}}$ with an open-loop motion primitive $\langle \mathbf{a}_1, ... \mathbf{a}_\tau \rangle$, where $\tau$ was the fixed duration and $\mathbf{a}_i$, with $i\in[1, \tau]$, was the joint-space command sent to the robot at each robot timestep.
The design of these motion primitives was inspired by observing the behavior of standard quadruped robots and animals during locomotion.
Specifically, our motion primitives were defined as follows.
For a hip joint, we used a linear ramp.
For a knee joint, we use a half-sine with amplitude 1 or -1, corresponding to ``lift leg'' versus ``push leg into the ground'' primitives. A diagram of these motion primitives is shown in~\Cref{sec:motion_primitives_sarsa}.

We approximated the true action-value function $Q(\mathbf{s}, \bar{\mathbf{a}})$, where $\mathbf{s}\in\mathcal{S}$ was the state, using linear function approximation with a tile coder~\citep{sutton2018reinforcement}. Tile coding is based on the Cerebellar Model Articulator Controller (CMAC)~\citep{tilecoding_james_albus} and was applied in RL before in \cite{watkins_thesis} and \cite{NIPS1995_8f1d4362}.
The robot's observation vector served as the agent's state.
Each state $\mathbf{s}$ was mapped by the tile coder to a sparse binary feature vector $\phi(\mathbf{s})$, and each agent action $\bar{\mathbf{a}}$ had a separate weight vector $\mathbf{w}_\mathbf{\bar{\mathbf{a}}}$. The estimated action value was defined by $\hat{Q}(\mathbf{s}, \mathbf{\bar{\mathbf{a}}})=\mathbf{w}_\mathbf{\bar{\mathbf{a}}}^{\top} \phi(\mathbf{s})$. 
Each agent action was selected using an $\varepsilon$-greedy policy with respect to the current action-value estimate.



Using the abstract agent interface, we applied the standard SARSA($\lambda$) algorithm with eligibility traces as described in \cite{sutton2018reinforcement}.
Because our tasks were non-episodic, learning proceeds without episodic resets, and eligibility traces decay without being cleared at episode boundaries. 
With this formulation, SARSA($\lambda$) operated as a fully online, continuing-control algorithm directly on the physical platform.
The results of this implementation are presented in~\Cref{sec:sarsa_results}.


\subsection{Continuous Action RL: Soft Actor-Critic (SAC)}

Soft Actor-Critic (SAC) is an off-policy RL algorithm designed for continuous action spaces.
SAC combines actor–critic methods with entropy regularization.
The original algorithm~\citep{sac} learns a stochastic policy network, two action-value functions, and one state-value function network, all using a replay buffer.
Newer implementations of the SAC algorithm omit learning the state-value function and achieve similar performance~\citep{SpinningUp2018}.
We have modified the CleanRL implementation~\citep{huang2022cleanrl} and applied it to the Physical Ant, with the observations and actions presented in~\Cref{sec:obs}.
\revision{The action bounds were adjusted to constrain the knee to a $20^\circ$ range with a $50^\circ$ offset, and the hip to a $45^\circ$ range with $0^\circ$ offset.}




\section{Reinforcement Learning Results}\label{sec:results}

We begin by introducing the common methodology employed by both SARSA($\lambda$) and SAC, where learning is performed directly on board the Physical Ants.
We then present the hardware results of these two reinforcement learning algorithms introduced in~\Cref{sec:benchmarking_RL}.

\paragraph{Common Methodology.}
The task was the back-and-forth objective defined in~\autoref{eq:back_and_forth_reward}, in which the robot must walk within a specified circle.
For all experiments, the radius of this circle was set to $R = 0.3$~m.
This value was determined by the dimensions of the operational arena
(\Cref{fig:system_architecture}(a)).
At the start of each trial, the power and communication cables were positioned outside the circular region. 
The Physical Ant was then placed inside the circle and the cables (power and communication) were arranged so that they were not entangled with each other and with the robot's body. 
We refer to this configuration as the start configuration.
During learning, the robot occasionally became entangled in the cables.
When this occurred, the experimenter manually paused the experiment, \revision{disentangled the robot}, then reset the robot by returning it to the start configuration \revision{within the circle}, and resumed the experiment.
We have also ensured that the interaction was performed within the same fixed-duration control loop $\Delta t$. \revision{Lastly, for both tasks, have not designed a complex reward for the experiments presented, but used a simple progress reward, as seen in Equation~\ref{eq:back_and_forth_reward}.}

Each RL algorithm was run on the robot for 80 minutes per trial, and we conducted five independent trials.
SARSA($\lambda)$ was evaluated only on the Physical Ant, and
SAC was evaluated on both the Physical Ant and the Physical Ant~Lite. 
The evaluations used the performance metric from~\Cref{sec:measuring_performance}.

\paragraph{Evaluation of SARSA($\lambda$).}\label{sec:sarsa_results}

\begin{figure}[t]
    \centering
    \includegraphics[width=\linewidth]{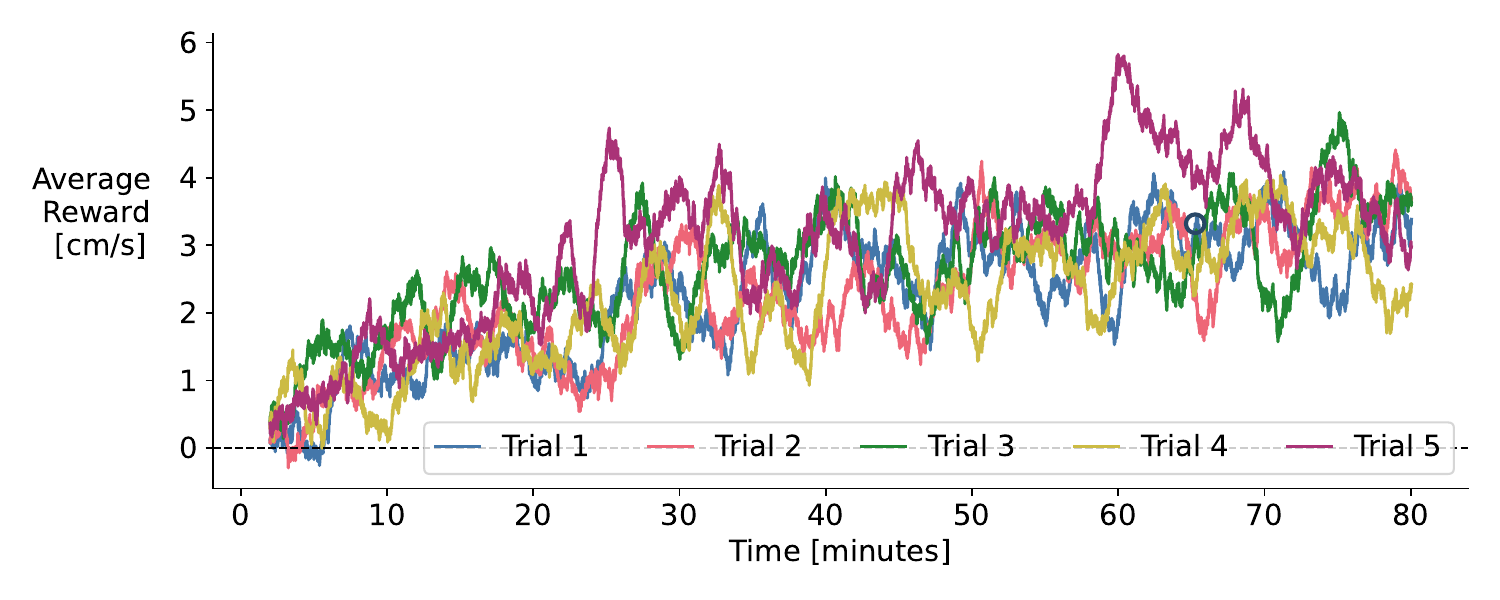}
    \caption{\textbf{Run-time learning performance of SARSA$(\lambda)$ executed onboard the Physical Ant across 5 trials.} Average reward is computed using~\autoref{eq:average_reward} on a window of 120 seconds.
    These 5 trials had a total of one interruption (indicated with circles) \revision{caused by the leg failure in~\Cref{fig:failures}(g). The leg was repaired within 10 minutes and the experiment was continued.}
    The experiments were run on a Desktop computer with the Intel Core i9-13900K CPU.
    }
\label{fig:sarsa_back_and_forth}
\end{figure}

Learning with SARSA($\lambda$) on hardware included several design decisions.
First, to select good algorithm parameters at design time, we used Optuna~\citep{optuna} for automated parameter optimization with the simulation.
We conducted a sweep over the parameters listed in~\Cref{sec:param_sweep_sarsa_lambda}, evaluating each of 100 configurations across \revision{ten} random seeds. The configuration that achieved the highest average reward was selected for the final experiments and is shown in~\Cref{tab:sarsa_parameters}.
\begin{table}[H]
\caption{Parameters for SARSA($\lambda$) with tile coding used for the experiments in~\Cref{sec:sarsa_results}.
    }
    \centering
    \begin{tabular}{c|c|c|c|c|c|c|c|c}
    \makecell{ $\Delta t$ }  &
          \makecell{ Elig. Tr. \\ $\lambda$ }
           & $\gamma$ & $\epsilon$ & Step Size & \makecell{ Tiles per \\ Dim.}  & Tilings & \makecell{ Tile Coder \\ Table Size}  & \makecell{  Actions} \\
        \hline
     0.5 s & 0.964 & 0.998 & 0.255 & 0.0001 & 4 & 192 & $2^{25}$ & 8 \\
    \end{tabular}
    \label{tab:sarsa_parameters}
\end{table}

Next, the specification of observation bounds for the tile coding was important. 
Because we used a tile coding library \href{http://incompleteideas.net/tiles/tiles3.html}{Tiles3} that depends on the scale of the inputs, we precomputed realistic observation ranges by driving the robot through diverse motions in multiple directions.
These measured bounds were then used to normalize the observations prior to the generation of tile-features by the software library.

Third, the design of the motion primitives and ensuring they are realizable on hardware substantially influenced learning. 
In particular, we found that primitives coordinating two legs simultaneously, rather than actuating a single leg in isolation, produced more dynamically meaningful behaviors.

Results from five independent trials are shown in~\Cref{fig:sarsa_back_and_forth}. 
In every trial, the agent achieved slow yet competent walking behaviors (average reward of 2-4 cm/s) within approximately one hour of on-hardware learning. The key result is that learning is reliably observed across trials, even with a very simple RL algorithm.

\paragraph{Evaluation of SAC.}\label{sec :sac_results}

We implement Soft Actor–Critic (SAC) on the two Ant platforms, the Physical Ant and the Physical Ant Lite. Both platforms share identical observation spaces, action spaces, and reward functions.
 Except for the modifications stated below, the algorithm parameters (\Cref{tab:sac_parameters}) and network architectures for the policy and action-value functions follow the default configuration in the implementation of CleanRL~\citep{huang2022cleanrl}.

First, we reduced the number of random interaction steps before learning begins from 5000 to 2000 (approximately 4 minutes of on-hardware random interaction). Empirically, this earlier start of gradient updates produced average reward performance comparable to the default setting, while reducing random actions taken on the robot, which could potentially damage it, and enabling faster learning (see~\Cref{sec:effect_of_applying_layer_norm_reward_scaling_learning_starts} in the supplementary material).

Second, we incorporated Layer Normalization (LayerNorm)~\citep{ba2016layernormalization} into both the policy and critic networks. 
Prior work suggests that LayerNorm can stabilize learning~\citep{elsayed2024streamingdeepreinforcementlearning}; we verified this effect using our Simulated Ant, as shown in the supplementary material~\Cref{sec:effect_of_applying_layer_norm_reward_scaling_learning_starts}, across 30 simulation seeds. LayerNorm both  accelerates learning and reduces variance. This faster convergence is particularly important in hardware settings, where interaction time is costly.

Third, we found reward scaling to be important for stable SAC training. 
The SAC algorithm has an entropy component that is sensitive to the reward scale. 
An initial reward scaling factor of $1.0$ led to poor learning performance on hardware, whereas scaling rewards by a \revision{larger factor} substantially improved performance and enabled consistent learning across seeds. 
We validated the effect of the scaling reward parameter in simulation across $30$ seeds (supplementary material~\Cref{sec:effect_of_applying_layer_norm_reward_scaling_learning_starts}) and demonstrated that a larger value produces faster learning. 

\revision{Lastly, we apply the modifications recommended in~\cite{deasis2024idiosyncrasytimediscretizationreinforcementlearning} to the return definition. This alleviates a idiosyncratic dependence on time-discretization for the algorithm. It decouples the optimization objective from $\Delta t$, making it a solution parameter instead of a problem parameter.}

\begin{figure}[t]
    \centering
    \includegraphics[width=\linewidth]{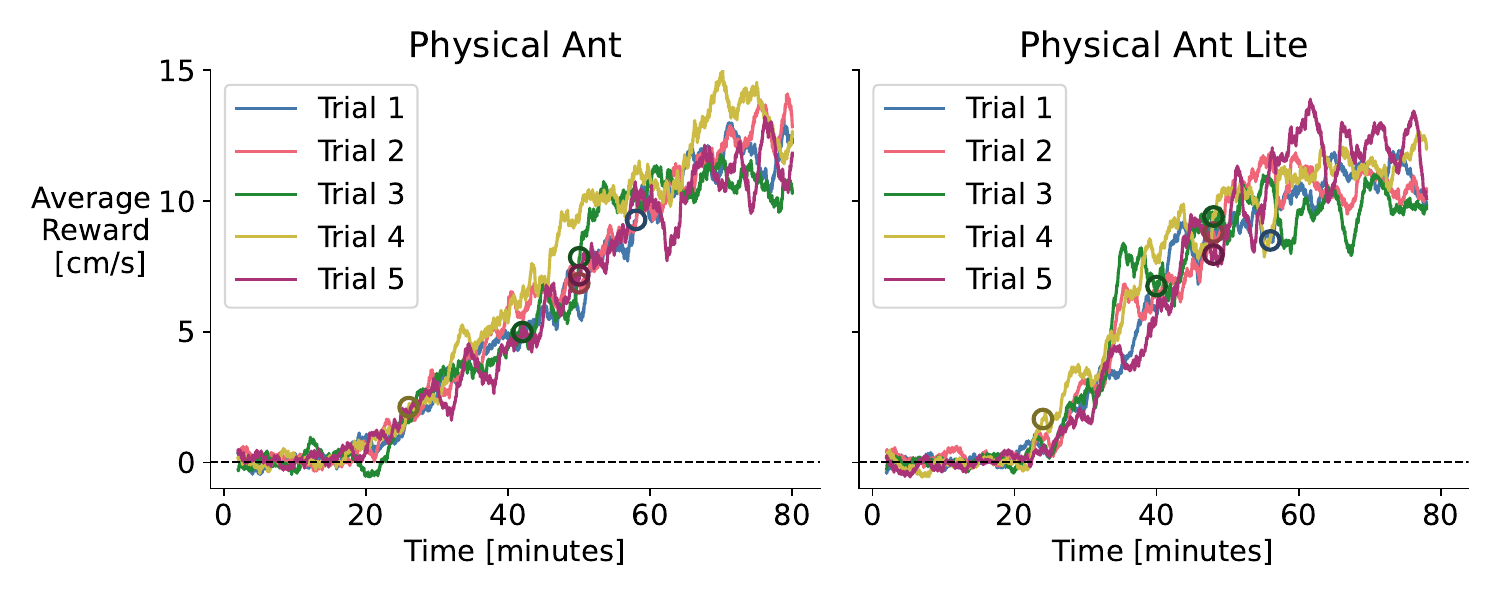}
    \caption{\textbf{Run-time learning performance using SAC on hardware for 5 trials.} Average reward is computed using~\autoref{eq:average_reward} for both the Physical Ant and the Physical Ant Lite on a window of 120 seconds. Circles indicate interventions, which are manual stops due to cable entanglement with the robot's legs.
    The experiment was run on a Macbook M1 \revision{with 16 GB of memory. The performance for one of the runs can be seen in \href{https://www.youtube.com/watch?v=QJuqB7gb4y0&t=322s}{Movie 2}}.}
    \label{fig:performance_learning}
\end{figure}

\Cref{fig:performance_learning} shows the average back-and-forth reward [cm/s] over time for five trials, on both physical platforms. By the end of training, all trials on both platforms exhibit stable, visually plausible walking behaviors, demonstrating that SAC can reliably learn locomotion directly on hardware in this configuration.  We note that the SAC behavior had more interventions to clear cable entanglement than encountered with SARSA($\lambda$), but the robot was also moving more with the SAC policies. 

\begin{table}[t]
    \caption{Parameters for SAC used for the run-time learning on the Physical Ants.}
    \centering
    \begin{tabular}{c|c|c|c|c|c}
        $\gamma$ & $\tau$ & \makecell{Batch  Size} & \makecell{Buffer  Size} & \makecell{Learn  Start} & \makecell{Reward  Scale} \\
        \hline
        0.99 & 0.005 & 256 & $10^{6}$ & 2000 & 100.0 \\
        \hline \hline
       $\Delta t$ &  \makecell{ Actor \\ Step Size } & \makecell{Critic \\ Step Size } & \makecell{Entropy Coeff. \\ Step  Size } 
        & \makecell{Policy Update \\ Frequency } 
        & \makecell{Target Update \\ Frequency } \\
        \hline 
        0.12 & 0.0003 & 0.001 & 0.001 & 2 & 1  \\
    \end{tabular}    \label{tab:sac_parameters}
\end{table}

\paragraph{Translating policies from the simulator to physical reality using SAC.}\label{sec:sim_to_real}


A popular approach to obtain competent robot behavior is to simulate an approximation of the robot and its environment, and learn a policy in simulation.
The policy can then be deployed on the real robot as is, or adapted on the robot with additional experience. 
\begin{wrapfigure}[17]{R}{0.50\linewidth}
    \centering
    \vspace{-0.5cm}
    \includegraphics[width=\linewidth]{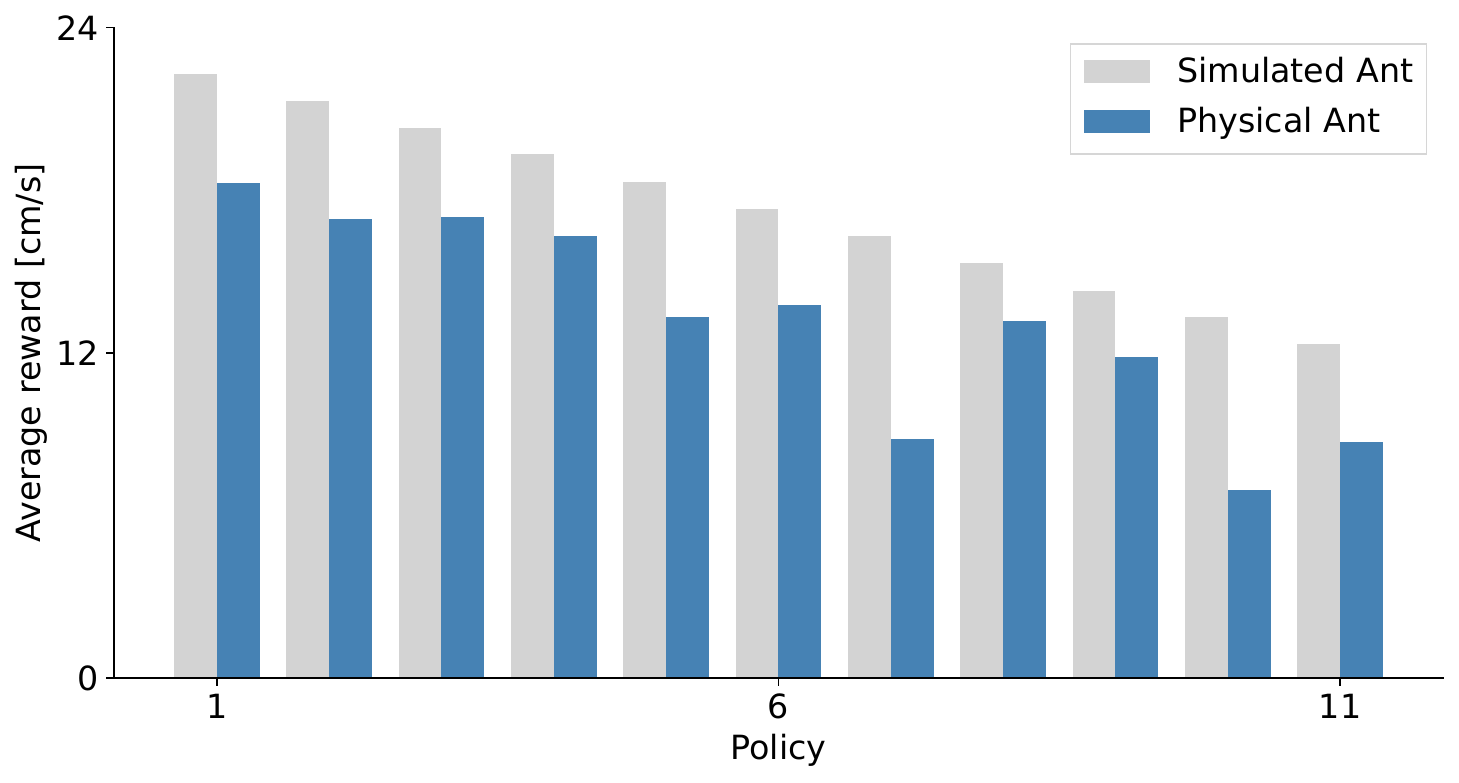}
    \caption{Evaluating policies learned using SAC on the Simulated Ant with the Physical Ant on the back-and-forth task. We chose ten policies of the Simulated Ant with reward rates between \revision{23} cm/s to \revision{13} cm/s and measured their performance on the Physical Ant.
    }
    \label{fig:sim2real}
\end{wrapfigure}

We tested how policies learned on the Simulated Ant performed on the Physical Ant \revision{without further learning}. 
We learned 2,304 policies in simulation using SAC with several parameter configurations (supplementary material~\Cref{sec:parameter_sweep}) for a duration of 400,000 timesteps (the equivalent of about 13 hours). We have not done any domain randomization~\citep{domain_randomization}. 
We then sampled ten policies, such that the \revision{average reward} of the policies in simulation were uniformly distributed between \revision{13} cm/s to \revision{23} cm/s (which was the fastest policy observed). We label these policies 1 to \revision{11} in decreasing order of their performance in simulation, with policy 1 being the best and policy \revision{11} being the worst. Finally, we evaluated these policies on the Physical Ant for \revision{5,000} steps (\revision{10} minutes), with no interventions. We  report the mean performance in~\Cref{fig:sim2real}. 
We noticed two key results.

First, all policies transferred with some success and enabled the robot to walk in the correct direction.
This shows that the Simulated Ant captures the key aspects of the Physical Ant well.
We also noticed that the less proficient policies make the ant entangled with the cable a lot more, \revision{for example policy 10 and policy 7}.

Second, the ranking of the policies changed when they were transferred to the Physical Ant. For example, the policy that performed the best on the Simulated Ant was the sixth best on the Physical Ant. 
This change in ranking is an important result because policy improvement requires the ability to compare two policies. If the comparison is not accurate, then finding the best policy from a large number of policies learned in simulation poses challenges for hardware deployment.

Note that the policies from our sim-to-real analysis are not directly comparable because they are chosen from a set of parameters that do not always overlap with the parameters used for learning from physical experience (\Cref{tab:sac_parameters}).

\section{Examining the Platform Suitability for RL researchers}\label{sec:suitability_for_RL}

Most RL researchers have limited familiarity with robotics.  Indeed, the authors have frequently heard  RL researchers say that they avoid robotics experiments, due to difficulties they experienced in the past. One common problem is the long delay to the first successful experiment for a researcher using a novel robot with RL.  Another common problem is the difficulty in repairing or adapting the physical system when failures arise.  We share our experiences on both these common concerns. We have observed that multiple researchers could use the platform within a few days to learn policies, and we have found that many failures with the hardware can be rapidly corrected, due to the 3D printed design, and the use of commercially available (COTS) components.

\subsection{Easy for Many to Build and Use}

An important goal for the Open Ant platform is to enable researchers to build, deploy, and iterate on the platform with minimal effort.
To date, the robot has been independently assembled and tested across multiple locations, including Canada, USA, and Malaysia. 
Once all the components were purchased, it took between 2 to 5 hours to assemble a full robot.
The Open Ant was used for experimentation with RL at two meetings of academic researchers, with initial use at a workshop and a more thorough testing at the \revision{\href{https://www.openmindresearch.org/winterschool2026}{Openmind Research Institute Winter School} in Malaysia.}

The winter school provided evidence that the platform can be successfully used by researchers who were inexperienced in RL and robotics. At the school, five independent teams (total of 20 participants) successfully used and modified SAC and SARSA($\lambda$) implementations.  The participants tested sim-to-real transfer on the physical robot in an episodic setting (a walking forward task).  
The teams consisted of researchers who had limited experience with robotics and RL, and no prior experience with this particular platform. 
Nevertheless, all teams were able to demonstrate successful learning and transfer with three days of access to the physical platform.
The physical platform was assembled by a local team of university robotics \revision{students}, with a combination of locally available 3D printers, and externally sourced commercial components. 
\revision{The assembly process highlighted the platform's low barrier to entry; as noted in a testimony from the local assembly team: ``once we understand how all the parts fit together, it takes less than two hours to assemble everything. When we tried building it (the Open Ant) without a guide, it took us around four to six hours working on it on and off. Most of that time was spent assembling, realizing mistakes, and then taking things apart to fix them''}.
The participants used their own laptops to control the systems with a variety of operating systems (Linux, MacOS, Windows). \revision{An illustration of the experiments performed at the Winter School can be seen in \href{https://www.youtube.com/watch?v=w6y4_IwT_G0}{Movie 6}.}
These deployments demonstrate that the system is accessible to a broad range of researchers, and that the platform provides a rapid path to successful RL experiments with a robot.

\revision{We speculate that some design choices make this a good introductory physical platform for RL researchers.} The combination of the simplicity of the platform, AC-powered long-duration operation, a standard Gymnasium interface, and compatibility with external compute platforms created a streamlined and engaging researcher experience.

\subsection{Platform Reliability} \label{sec:Reliability}

\begin{figure}[t]
    \centering
    \includegraphics[width=0.6\linewidth]{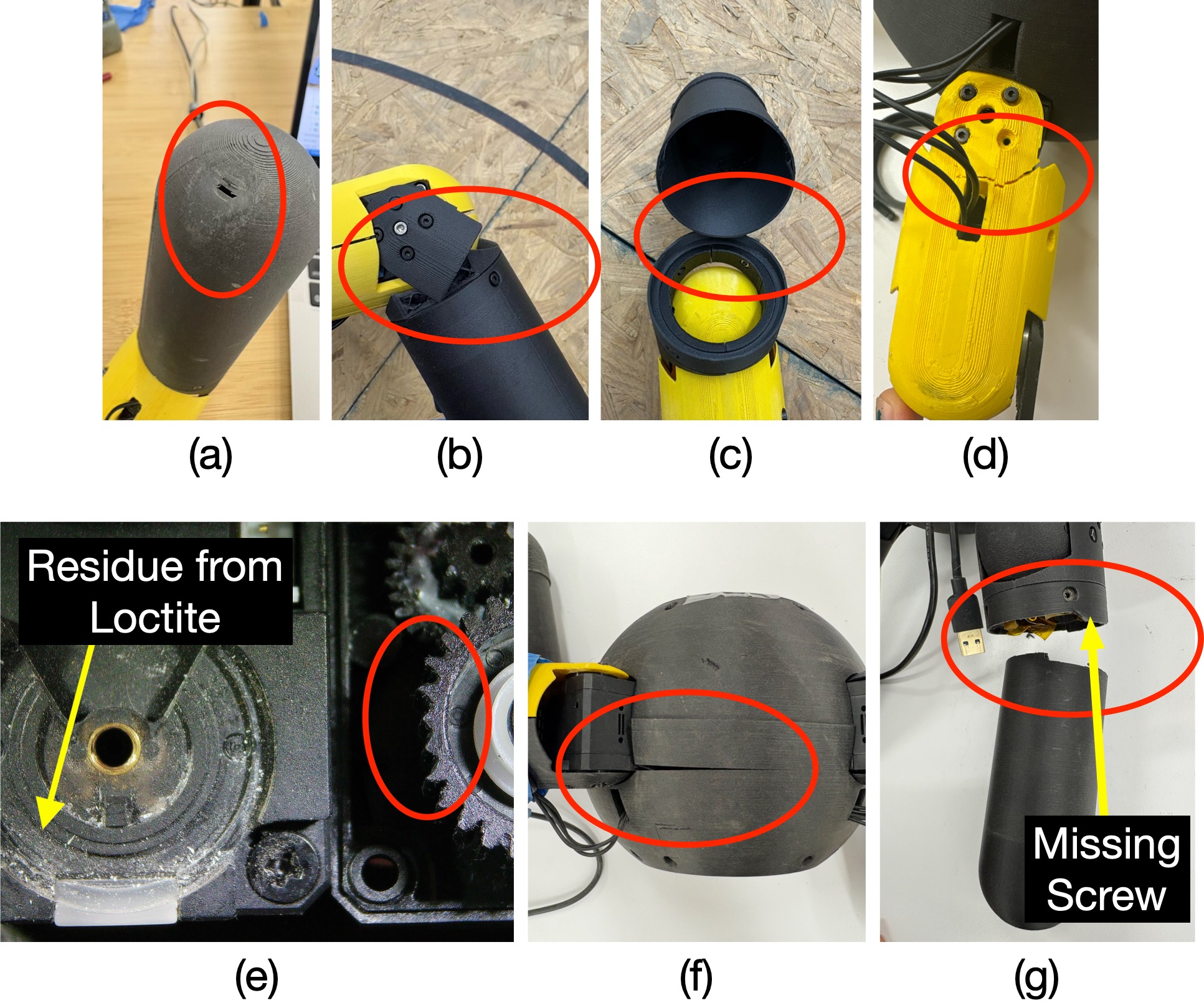}
    \caption{\textbf{Different failures observed during the development of the Physical Ant} (a) Foot-tip abrasion from repeated impacts, mitigated with 3D-printed TPU socks. (b) Hip cover fracture due to locomotion stress, redesigned with a stronger interface. (c) Leg shell damage after manual overloading; reinforced with internal ribs. (d) Leg housing cracks from accumulated stress, likely related to (e), interface strengthened. (e) Motor contamination from excess Loctite and plastic gear wear (XL430-W250-T). (f) Outer shell crack, mitigated by tightening screws to maintain structural integrity. \revision{(g) Leg broken during one of the SAC experiments. We notice a missing screw which could have affected the load distribution.}}
    \label{fig:failures}
\end{figure}

From earlier experiments performed during the development of this platform, we observed several mechanical failure modes on the Physical Ant. As shown in~\Cref{fig:failures}, these issues were primarily caused by ground impacts, joint loadings, and accumulated stress in 3D-printed components.
In response, we iteratively improved the hardware design by reinforcing interfaces, redesigned vulnerable parts, and added protective elements such as 3D printed Thermoplastic Polyurethane (TPU) foot socks. We have also performed thermal improvements based on a detailed thermal analysis (\Cref{sec:thermal_analysis}), and  power and communication fixes (\Cref{sec:electrical_analysis}).
These modifications were important for enabling long learning runs. By releasing the hardware design as open-source, we aim to encourage further community-driven improvements to the platform’s structural robustness and long-term reliability.

\section{Discussion and Future Directions}\label{sec:discussion_and_future_directions}

\revision{We have presented} the Open Ant as a platform for RL researchers who are familiar with simulation domains to include robot experiments in their evaluations.
For our primary purpose of easing adoption, we have made the platform intentionally close to existing simulation environment and tasks, so that researchers can succeed with their initial attempts.

An RL researcher who is familiar with algorithm development in \revision{simulation} may question the value of including robots as part of their research methodology.
One argument in favor is that if our ultimate goal is to build learning agents that operate in the physical world, then experiments with the algorithms in the physical world are essential.
Even for research programs that target non-physical domains, like agents for the PlayStation game GranTurismo~\citep{wurman2022outracing}, researchers may not have direct access to the deployment environment with human participants. Those researchers may also find it valuable to evaluate their algorithms with a physical robot, where they can access a non-simulated deployment environment.
A physical robot platform provides a valuable testbed for validating algorithms and exposes discrepancies between simulation and reality that would otherwise remain hidden.



\subsection{Practical Concerns for Future Research}

We consider some factors that can make the direct application of reinforcement learning algorithms to hardware more practical. These considerations are taken with respect to the current state-of-the-art.

\textbf{Random Exploration.} To safely deploy \textit{current} learning algorithms on physical robots, their typical random exploration must not compromise the integrity of the robot.
For example, motors, gearboxes, and structural components must tolerate the wear induced by this exploratory behavior.
The Open Ant is designed to tolerate such exploration; it uses relatively large motors for its size, and the joint limits and kinematics prevent any self-collisions. 
The safety requirement imposes constraints on the robot design or the algorithm. The Open Ant addresses this issue in a straightforward way by over-sizing its actuators. More complex robots could require substantially more effort.

\textbf{Unrecoverable States and Resets.} Second, unrecoverable states should be rare.
For example, our the robot can continue walking even if it rolls onto its back, provided the full knee range of motion ($\pm70^\circ$) is used.
The back-and-forth locomotion task was designed to require minimal human intervention by keeping the robot within the camera field of view.
This allows the robot to learn without frequent interventions. \revision{The cable entangling with the robot did occasionally require intervention and in the future we will explore mitigating this through the use of a cable management system utilizing elements such as retractable cables, slip rings, or pulley systems.}
An adaptation of the Gymnasium humanoid walking task, an alternative we considered early in the design phase would require interventions which are trivial to implement in simulation (by resetting the environment) but burdensome in physical experiments.
In summary, the design of the physical experiment should minimize physical human interventions and resets.

\textbf{Performance vs. Compute} We also observed that learning performance can vary depending on the specific computer instance on which the agent is executing (\Cref{sec:timing}).
Although we adjust the duration of the environment interaction step to maintain a fixed interaction frequency, the timing between issuing a motor command and reading sensor values for the next observation will depend on the time the agent takes to decide on its action.
The impact of this timing on the Open Ant platform, as well as other factors that depend on compute speed, like camera processing latency, deserve further analysis.
Furthermore, conventional agent environment interfaces, such as Gymnasium, lack specification of timing aspects that are unavoidable when operating in the real world. We acknowledge that timing remains a concern despite some existing works addressing it, for example~\cite{pmlr-v330-farrahi26a, yuan2022asynchronous, 10.1109/IROS.2018.8593894}.

\textbf{Environment Changes}. The environment changes over time as the agent learns from physical experience.
For example, we observed that the robot gradually created holes in its feet during learning, altering their structural integrity, as shown in~\Cref{fig:failures}(a). We have mitigated this by building 'socks' out of thermoplastic polyurethane material.
In addition, repeated walking on a Medium-density fiberboard (MDF) wooden floor generated fine dust, which accumulated over time and changed the friction between the feet and the surface.
Such effects are difficult to capture accurately in simulation and are often difficult for the designer to anticipate in advance, which motivates continual learning.

\textbf{Reward Drop}. 
We observed occasional runs in which the performance in simulation drops to nearly zero.
Similar behavior has also been observed on the physical robot, as illustrated in \href{https://www.youtube.com/watch?v=tHW0VfXyEPQ}{Movie 4} for SARSA($\lambda$). In both cases, the agent eventually recovers and resumes learning, as shown in \href{https://youtu.be/NCSODJeMJPw}{Movie 5} for the simulation.
It is unclear what precisely causes this performance drop, but it requires further investigation. This phenomenon is illustrated in~\Cref{fig:figure_reward_drop} in~\Cref{sec:reward_drops}.

\textbf{Simulator Exploits}. 
A practical limitation of sim-to-real is that policies can exploit inaccuracies in the simulator rather than learn behaviors that transfer to the real world. 
\revision{Workarounds include} intensive domain randomization~\citep{domain_randomization} \revision{and} reward engineering.
Although this concern does not apply to our hardware learning experiments, it was present in our early sim-to-real experiments (\Cref{sec:sim_to_real}). 
As discussed in~\Cref{sec:sys_id} and shown in \href{https://www.youtube.com/watch?v=9VMjJRwFiRw}{Movie 3}, we observed policies that achieved high rewards in simulation by rapidly jittering the robot's feet, resulting from a bang-bang control strategy.
While highly rewarded in simulation, this behavior could not be reproduced on the physical robot.
This example illustrates how policies may exploit imperfect dynamics or contact models that are absent in the real world.
Learning directly on hardware avoids this failure mode because optimization occurs under the true system dynamics rather than a simulated approximation.


\subsubsection*{Acknowledgments}
\label{sec:ack}
\revision{The authors would like to thank Richard S. Sutton, Jun Luo, and Arsalan Shariffnassab for engaging discussions, Mohamed Elsayed for the suggestions to use CleanRL and LayerNorm for the SAC implementation, and Jose Uribe for help with some of the experimental setup building.
We would also like to thank students Kevin Chew Ken Yi, Sim Sheng Wei and Lim Zi Quan at Universiti Tunku Abdul Rahman (UTAR), Malaysia, for implementing five instantiations of the Physical Ant for the \href{https://www.openmindresearch.org/winterschool2026}{Openmind Research Institute Winter School}.
We thank the NASA Jet Propulsion Laboratory for access to lab space and the Openmind Research Institute for funding.
}

\bibliography{main}
\bibliographystyle{rlj}

\beginSupplementaryMaterials

\section{System Identification}
\label{sec:sys_id}
\input{appendix/appendix_sys_id}
\cleardoublepage

\section{Torque Calculations}
\label{sec:torque_calculations}

\input{appendix/appendix_torque_calculations}

\cleardoublepage

\section{Failure Modes and Observed System Behaviors}
\label{sec:failure_modes}

\input{appendix/appendix_faults}
\cleardoublepage

\section{Parameter Sweep for Sim2Real Experiments}
\label{sec:parameter_sweep}
\input{appendix/appendix_parameter_sweep}
\cleardoublepage

\section{Parameter Sweep and Simulation Results for SARSA($\lambda$) Experiments}
\label{sec:param_sweep_sarsa_lambda}
\input{appendix/appendix_parameter_sweep_sarsa_lambda}

\section{Extra results}
\label{sec:extra_results}
\input{appendix/extra_results}

\input{appendix/appendix_motion_primitives}

\input{appendix/appendix_list_movies}

\end{document}

%% file: appendix/appendix_sys_id.tex
We have taken several steps to improve the fidelity of the Physical Ant simulation model while keeping the effort reasonable.

\paragraph{Motor Damping.}
We computed the motor's damping coefficient as the ratio between the stall torque and the no-load speed, following typical DC motor guidelines~\citep{brushdc}.
From the Dynamixel datasheet, this yields 0.235 $\mathrm{Nms/rad}$ for the XL430-W250-T and 0.852 $\mathrm{Nms/rad}$ for the XM430-W350-T. These values were added to the joint \textit{damping} parameter in the MuJoCo MJCF XML.  \revision{This parameter can vary depending on the particular settings in the actuator's controller, so a more thorough system identification might benefit from measuring this parameter empirically}.

\paragraph{Motor Stiffness (Kp).}
We empirically measured the joint stiffness, which is the restoring torque per unit of angular displacement, in other words, if we perturb the robot's joint by $\Delta \theta$, where $\theta$ is the angle, the restoring torque is $\tau = K_p \Delta \theta$. 
This scalar was measured as follows: we placed the ant with one leg on a kitchen scale and deflected the leg downward by $0.1 \mathrm{rad}$ increments. For each angle, we recorded the vertical reaction force and the corresponding displacement. 
Across five measurements, the average stiffness values were $19.499 \mathrm{Nm/rad}$ for the XL430-W250-T, and $18.638 \mathrm{Nm/rad}$ for the XM430-W350-T. 
We used these values as the $K_p$ gains of the motor position controller in MuJoCo.

\paragraph{Velocity Gain $K_v$ and Motor Delay.} We additionally added a $K_v$ term for the velocity component of the controller which we tune to approximately match the response on the Physical Ant. After inspecting sinusoidal and square-wave joint tracking experiments, we observed that the lower-cost XL430-W250-T motor exhibits a one-timestep actuation delay. This time delay was not included yet in the simulation environment.

\paragraph{Foot - Ground Friction.}
We empirically measured the sliding friction coefficient of the experimental floor by placing the ant on a tilted plate of the same material that the ground was made of. We obtained a friction coefficient of $\mu = 0.6$, which we used in the MuJoCo geometry friction parameters. The torsional and rolling friction components were kept very small because the robot’s feet have minimal rotational contact area and the legs do not undergo rolling motion, thus sliding friction dominates the contact mechanics.

\paragraph{Integrator Choice.} 
Finally, we used MuJoCo’s \textit{implicitfast} integrator, the recommended option from the MuJoCo documentation, as it provides a balance between numerical stability and computational performance.

The final parameters are in Table~\ref{tab:parameters_sys_id}. \revision{Note that the XC430-W240-T is missing from this table, but a similar analysis can be performed for that motor, as well.}

\begin{table}[H]
    \caption{Motor parameters from system identification.}
        \centering
        \begin{tabular}{c|c|c|c|c}
        \textbf{Motor} & \textbf{Kp} & \textbf{Kv} & \textbf{damping} & \textbf{max. torque at 12 V} \\ \hline 
        \textbf{XL430-W250-T} & 19.4 & 2.5 & 0.235 & 1.4 Nm  \\ \hline
        \textbf{XM430-W350-T} & 18.6 & 1.5 & 0.852 & 4.1 Nm   \\
        \end{tabular} \label{tab:parameters_sys_id}
\end{table}

\paragraph{Contact and Further Simulation Changes.} 

Once the physical parameters of the system (damping, stiffness, friction, and integrator settings) were set, we conducted a series of validation tests. 
In particular, we applied high-frequency motor commands and observed whether the robot exhibited unintended translational motion along the ground plane (see \href{https://www.youtube.com/watch?v=9VMjJRwFiRw}{Movie 3}).

This test is important for sim-to-real transfer: in our earlier (unreported) experiments, the RL policy exploited inaccuracies in the simulator by generating rapid, jerky leg motions that produced forward motion in simulation but not on hardware.
The result was high average reward in simulation despite a locomotion strategy that was physically unrealizable. 
We ensured that such high-frequency actuation does not cause the robot to ``slide'' in simulation by changing the \textit{solref} and \textit{solimp} parameters.
These two parameters control the softness and damping characteristics of MuJoCo’s contact model.

Lastly, after analyzing the effect of both the proportional gain $K_p$ and the derivative gain $K_v$, we observed that the resulting closed-loop behaviour becomes stiff, which means that the controller reacts too aggressively to small tracking errors, producing large corrective torque. This can lead to jerky motions. To compensate for this, we have added a time constant of $0.12$ seconds in the motor model of the Physical Ant simulation.

If researchers identify additional sim-to-real discrepancies, we encourage them to contribute improvements to the open-source project. Although our primary focus in this work is real-time learning on hardware, a high-fidelity simulation environment \revision{is} beneficial.

%% file: appendix/appendix_torque_calculations.tex
\begin{figure}[h]
    \centering
    \includegraphics[width=0.325\linewidth]{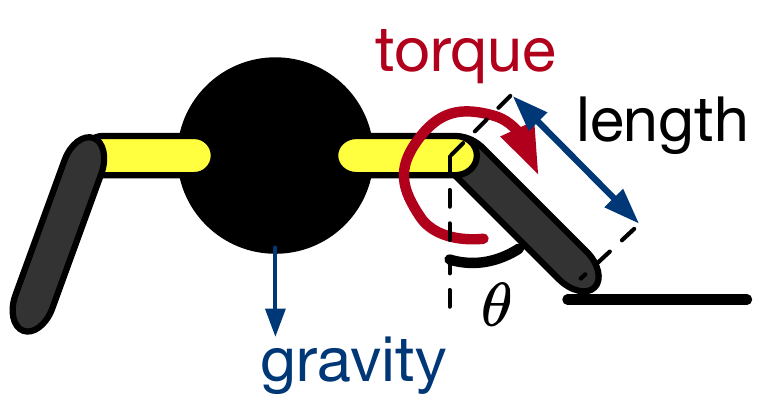}
    \caption{Free-body diagram of a single leg in stance for the torque calculations.}
\label{fig:force_diagram}
\end{figure}

In this section, we justify the selection of the Dynamixel actuators by estimating the maximum torque required at the knee joint during stance. 
Let the robot's total weight be distributed evenly among the $n_s$ legs in contact with the ground. 
The torque at the knee actuator is therefore
\begin{equation}\label{eq:torque_motor}
\tau_{\mathrm{motor}}=\frac{m g}{n_s} L \sin \theta,
\end{equation}
where $m$ is the total robot mass, $g$ is the gravitational acceleration, $n_s$ is the number of legs in stance, $L$ is the distance from the knee joint to the ground contact point, and $\theta$ is the leg angle relative to the vertical, as illustrated in~\Cref{fig:force_diagram}.

\paragraph{Physical Ant}
With a total mass of $1.2$kg, two legs in stance ($n_s = 2$), and a leg angle of $45^\circ$, Equation \ref{eq:torque_motor} yields a required torque of approximately $0.66$ Nm. 
The nominal torque of the XM430-W350-T actuator is $0.82$Nm, which slightly exceeds the required value with margin.
Therefore, this motor is appropriately sized for the Physical Ant and can sustain continuous learning in this setting.


In contrast, for the Physical Ant with plastic-gear motors \revision{(\textbf{Physical Ant Lite})}, the motors are underpowered.
With a total mass of $0.98$ kg, two legs in stance, and a leg angle of $45^\circ$, the required torque is approximately $0.54$ Nm.
The XL430-W250-T actuator provides a nominal torque of $0.28$ Nm, which is below the estimated requirement.
As a result, the lighter platform operates beyond the actuator's continuous torque capability during locomotion, leading to increased overheating, as discussed in~\Cref{sec:thermal_analysis}.

%% file: appendix/appendix_faults.tex
\subsection{Thermal Analysis}\label{sec:thermal_analysis}

\begin{figure}[H]
    \centering
    \includegraphics[width=\linewidth]{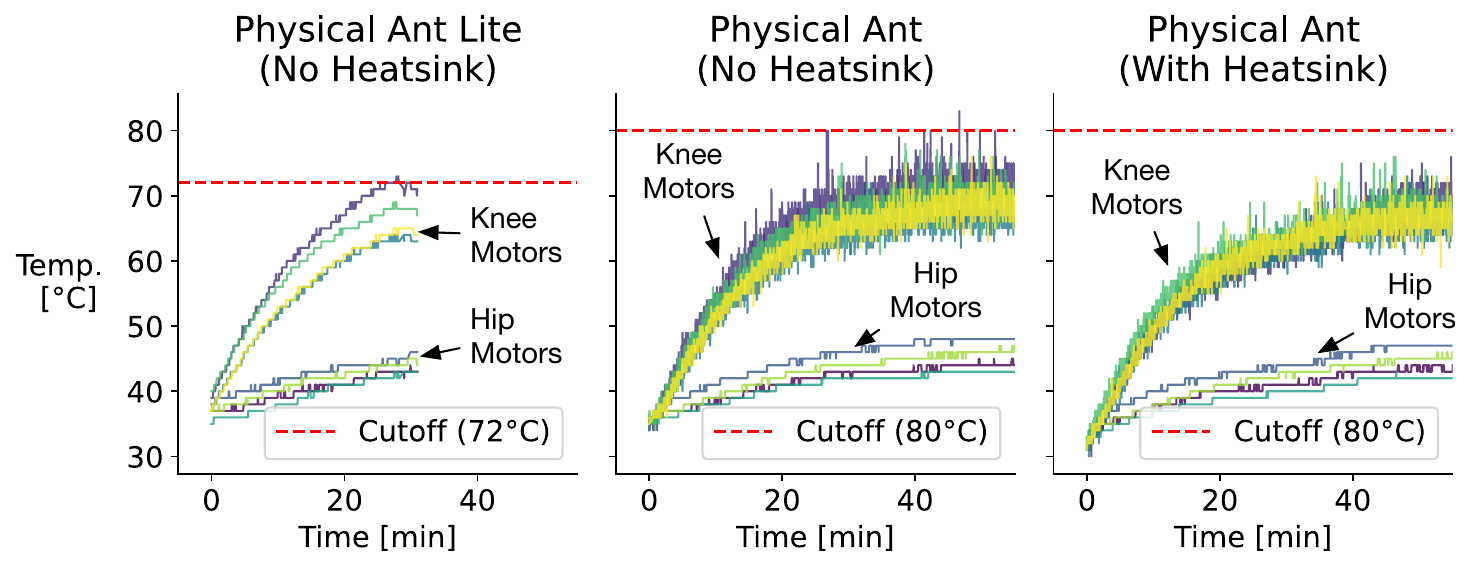}
    \caption{\textbf{Results for the thermal analysis during stress tests experiments}. Motor temperature over time during a one-hour stalled-load experiment for the Physical Ant Lite (XL-series motors) and Physical Ant (XM-series motors). The Physical Ant Lite platform reaches the 72$^\circ$ C thermal cutoff and shuts down after approximately 30 minutes, while the Physical Ant platform tolerates higher temperatures (80$^\circ$C limit). Integration of a heat sink on the Physical Ant reduces temperature spikes and prevents exceeding the thermal threshold during sustained operation.}
    \label{fig:thermal_heatsink}
\end{figure}

During early experiments with the Physical Ant Lite, we observed ``Overheating warnings'' repeatedly, in the status of the knee motors.
To better understand the thermal limits of the actuators, we conducted a stress test in which a heavy load was placed on the robot to mechanically stall the joints while continuously commanding a downward torque. Motor temperatures were monitored for one hour; the results are shown in~\Cref{fig:thermal_heatsink}.

Under these extreme conditions, one of the XL-series knee motors on the Physical Ant Light shut down after approximately 30 minutes, reaching its thermal cutoff at 72°C (the manufacturer-specified limit). In contrast, the higher-torque Physical Ant, equipped with XM-series motors, performed better. 
Although temperature spikes were observed, the motors remained operational up to their higher cutoff threshold of 80$^\circ$C.

\begin{figure}[H]
    \centering
    \includegraphics[width=0.4\linewidth]{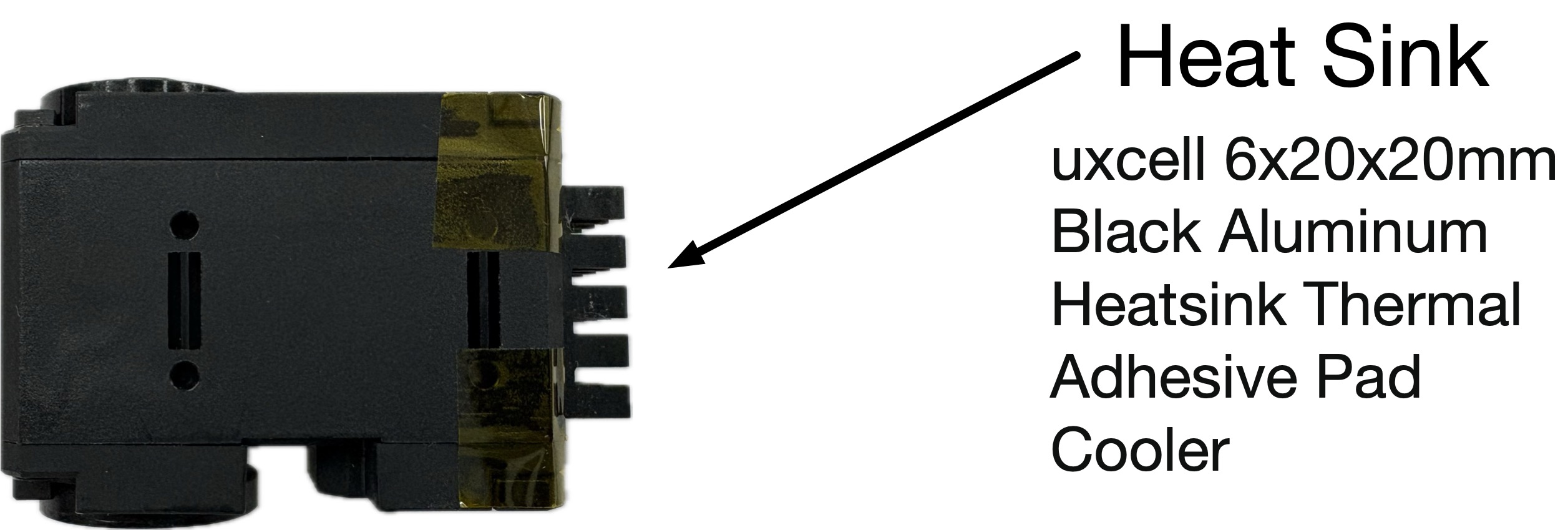}
    \caption{Motor equipped with heat sink.}
    \label{fig:heat_sink}
\end{figure}

We initially attempted to add passive heat sinks to the Physical Ant Lite, but observed minimal improvement likely due to the insulating properties of the plastic housing.
However, integrating heat sinks on the Physical Ant, \revision{like in~\Cref{fig:heat_sink}}, proved more effective: temperature spikes were reduced and remained below the 80$^\circ$C cutoff.

\subsection{Electrical Analysis}\label{sec:electrical_analysis}

In this section, we present an analysis of the intermittent communication dropouts observed during run-time learning, as well as other electrical considerations for the Physical Ants.

\begin{figure}[H]
    \centering
    \includegraphics[width=\linewidth]{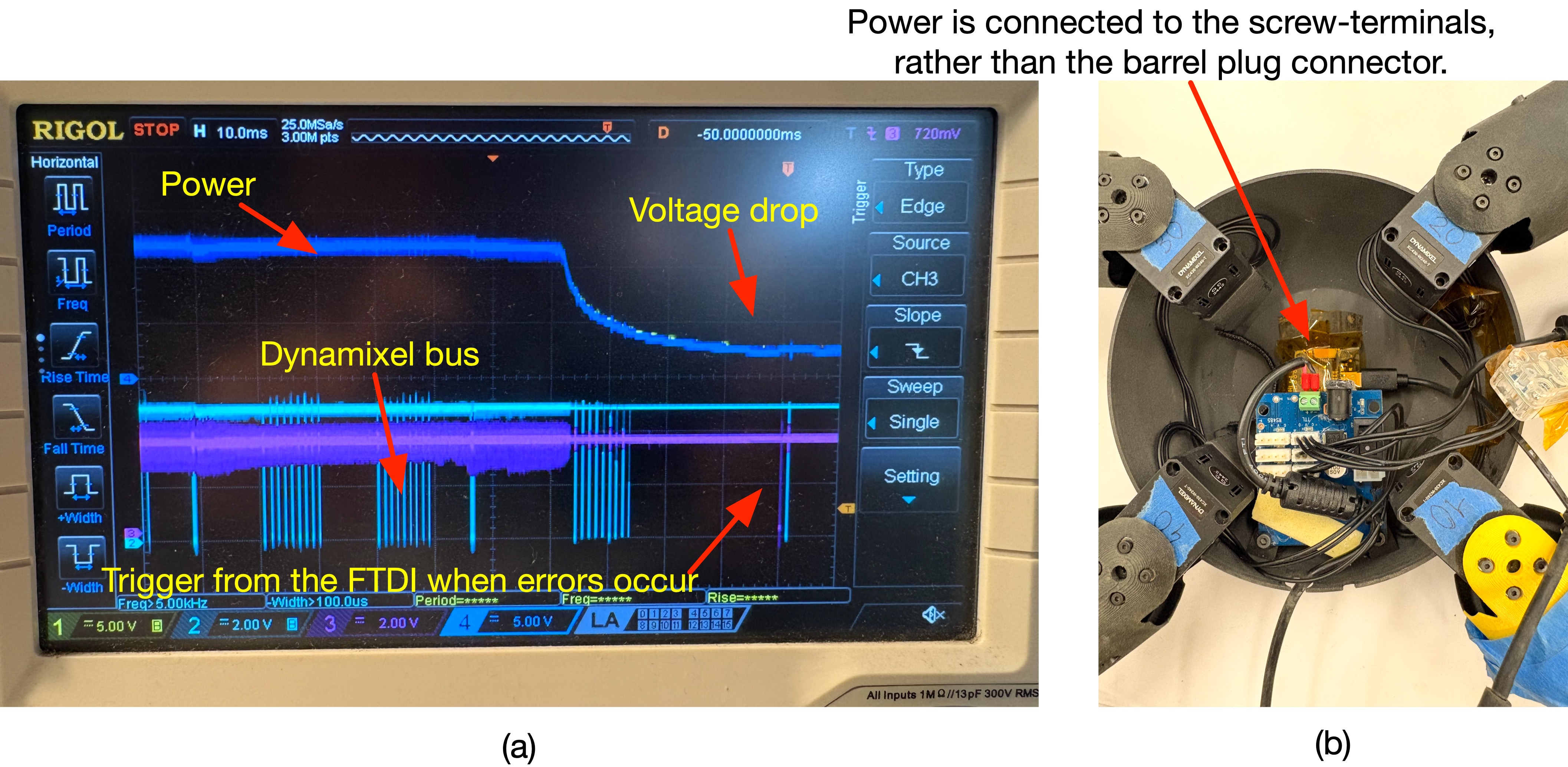}
    \caption{{\textbf{Electrical investigation of intermittent communication dropouts.} 
(a) Oscilloscope measurement showing the Dynamixel communication bus and supply voltage during a failure event. The oscilloscope was externally triggered by an FTDI USB-to-TTL converter when communication errors occurred. We observed that each error coincides with a brief drop in the supply voltage. 
(b) The power connection was modified to connect directly to the screw terminals on the control board rather than using the barrel connector. This eliminated the voltage drops caused by vibration-induced intermittent contact during learning experiments.}}
    \label{fig:oscilloscope_power}
\end{figure}

During long duration run-time learning experiments, we observed intermittent, very short communication dropouts that prevented fully autonomous operation of the robot.
To diagnose this issue, we used an oscilloscope to monitor the electrical signals out of the Dynamixel XM motors. The motor cable consists of three lines: signal, ground, and power.
Using an external triggered derived from an FTDI USB-to-TTL serial converter, we configured the oscilloscope to trigger when the communication errors occurred (see~\Cref{fig:oscilloscope_power}(a)). 
The measurements showed that each communication dropout coincides with a brief drop in supply voltage. Further inspection identified the root cause to be the barrel power connector, which we hypothesized can lose contact under the vibrations induced by learning.
Replacing the barrel power connector with a direct connection to the board's power terminal (as seen in~\Cref{fig:oscilloscope_power}(b)) eliminated the voltage drop.
We report this finding to assist other researchers who may encounter similar hardware-level issues during long-duration learning experiments.

%% file: appendix/appendix_parameter_sweep.tex
\begin{table}[H]
    \caption{Parameter sweep for learning policies on the Simulated Ant. We trained 6,912 policies in simulation, with the different parameter configurations given below. We picked 11 policies for evaluation on the Physical Ant.}
    \centering
    \begin{tabular}{c|c}
    \textbf{Parameter} & \textbf{Value(s)} \\ \hline
    Seed & 0, 1 \\ \hline
    Buffer Size & 1,000,000 \\ \hline
    $\gamma$ & 0.98, 0.92 \\ \hline
    $\tau$ & 0.005, 0.008 \\ \hline
    Batch Size & 128 \\ \hline
    Learning Start & 2000 \\ \hline
    Policy Step Size & 0.003, 0.001, 0.0003, 0.0001 \\ \hline
    Critic Step Size & 0.003, 0.001, 0.0001 \\ \hline
    Policy Update Frequency & 1, 2 \\ \hline
    Target Update Frequency & 1 \\ \hline
    Entropy Coefficient & 0.2 \\ \hline
    Autotune & True, False \\ \hline
    $\Delta t$ & 0.12 \\ \hline
    Number of environments & 1 \\
    \end{tabular}
    \label{tab:camera_ready_hyperparameters}
\end{table}

%% file: appendix/appendix_parameter_sweep_sarsa_lambda.tex
To select suitable parameters for SARSA($\lambda$) with tile coding, we used the parameter optimization framework Optuna~\citep{optuna}. 
Optuna performs automated parameter search by sampling configurations and evaluating them with respect to a user-defined objective.
In our experiments, the objective was \revision{to maximize} the average reward per second achieved by the agent during learning in simulation \revision{across 10 seeds}.
The search was conducted using the Tree-structured Parzen Estimator (TPE) sampler, a Bayesian optimization method, which is the default search method for the library.

\begin{table}[h]
    \caption{Parameter search space used for SARSA($\lambda$) with tile coding on the Simulated Ant. The parameters were optimized using Optuna with a Tree-structured Parzen Estimator (TPE) sampler. Each configuration was evaluated across 10 seeds. The notation ``[ ]'' shows an interval for reals and ``\{ \}'' shows an interval for integers.
    }
    \centering
    \begin{tabular}{c|c}
    \textbf{Parameter} & \textbf{Value(s)} \\ \hline
    Seed & 0, 1, 2, 3, 4, \revision{5, 6, 7, 8, 9} \\ \hline
    Number of Optuna Trials & 100 \\ \hline
    $\epsilon$ & $[0.01, 0.3]$ (log-scale) \\ \hline
    $\gamma$ & $[0.9, 0.999]$ \\ \hline
    $\lambda$ (eligibility trace) & $[0.0, 0.99]$ \\ \hline
    Tiles needed to cover an Observed Range & $\{4,\ldots, 16\}$ \\ \hline
    Number of Tilings & $\{2,\dots, 8\} \times$ 24 \\ \hline
    Step Size Base & $[0.01, 0.5]$ (log-scale) \\ \hline
    Reward Scaling & $[1, 10]$ (log-scale) \\ \hline
    $\Delta t$ & 0.05 s \\  \hline
    \revision{Motion Primitive Duration}  & \revision{0.5 s} \\

    \end{tabular}
    \label{tab:sarsa_optuna_hyperparameters}
\end{table}

\begin{figure}[h]
    \centering
    \includegraphics[width=\linewidth]{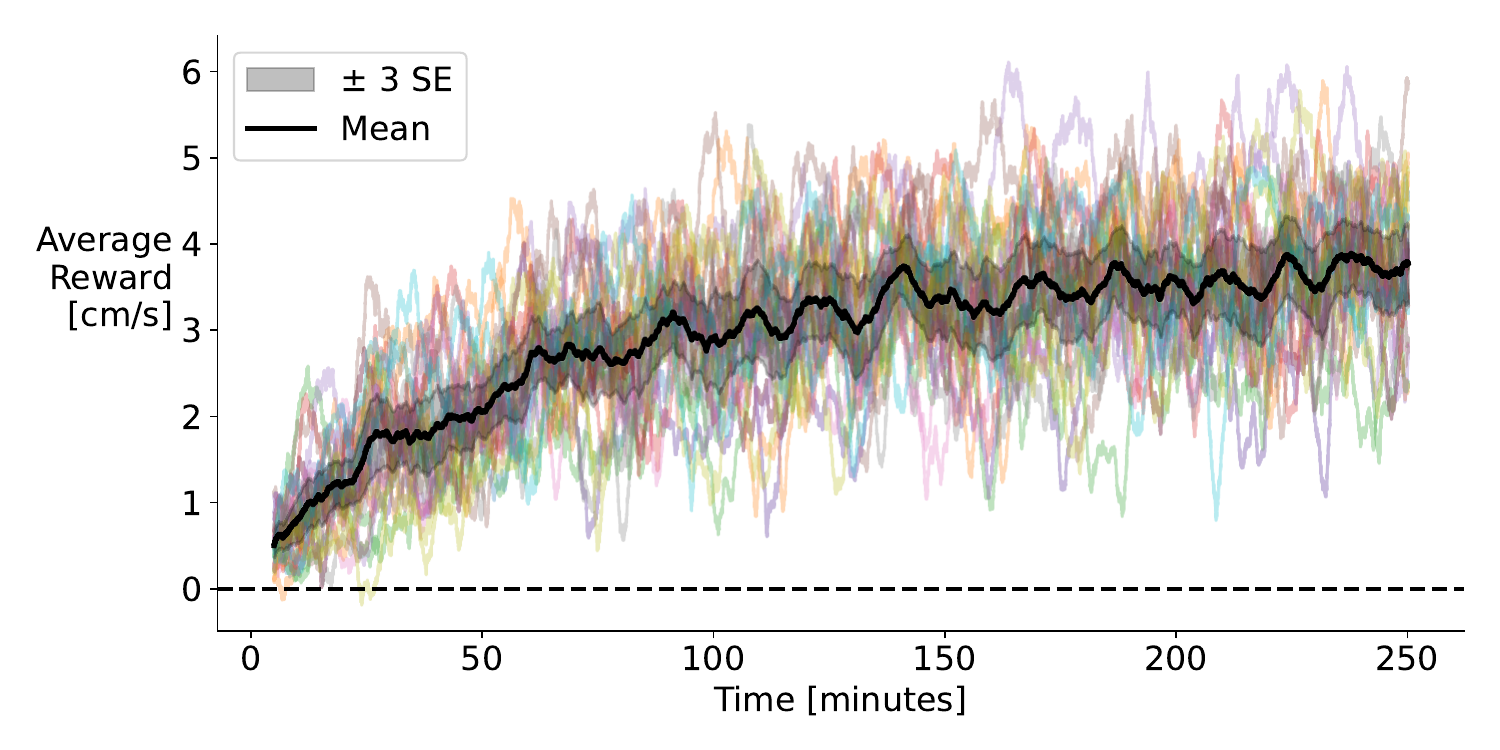}
    \caption{\revision{\textbf{Simulation results for SARSA($\lambda$)}. Using the parameters optimized by Optuna, this plot shows the mean performance with $\pm 3$ standard errors over 30 random seeds.}}
    \label{fig:simulation_results_sarsa}
\end{figure}

%% file: appendix/extra_results.tex
\subsection{Effect of Applying Layer Normalization, Changing when the Learning Starts, and the Reward Scaling in SAC}\label{sec:effect_of_applying_layer_norm_reward_scaling_learning_starts}

\begin{figure}[H]
    \centering
    \includegraphics[width=\linewidth]{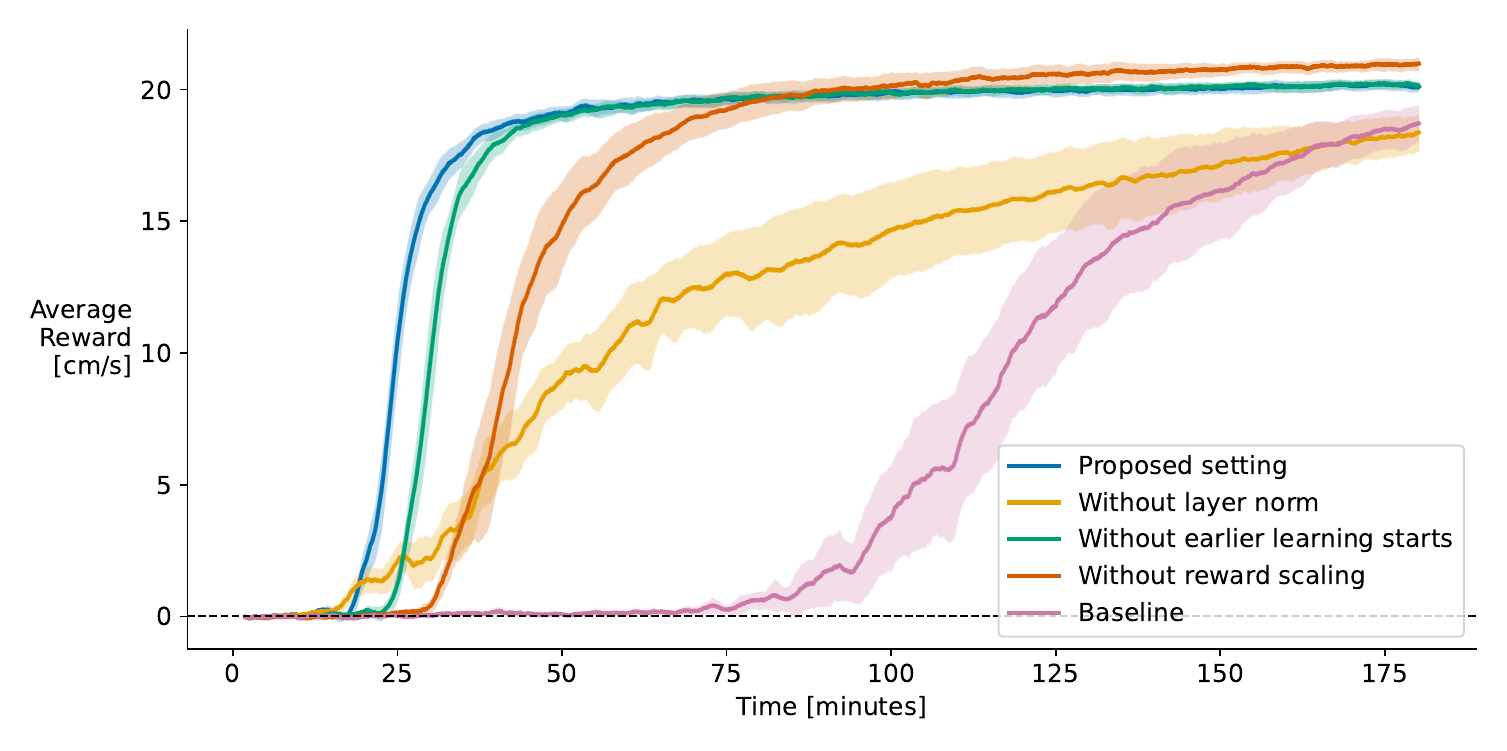}
    \caption{\textbf{Ablation on three of the components of SAC}. Simulation results showing the mean performance with $\pm 3$ standard errors over 30 random seeds for evaluating the effect of applying Layer Normalization for the policy and the critic networks, changing when Learning Starts parameter, and the Reward Scaling.}
\label{fig:layer_norm_simulation}
\end{figure}

\cleardoublepage
\subsection{Average Reward Drops}\label{sec:reward_drops}

\begin{figure}[H]
    \centering
    \includegraphics[width=\linewidth]{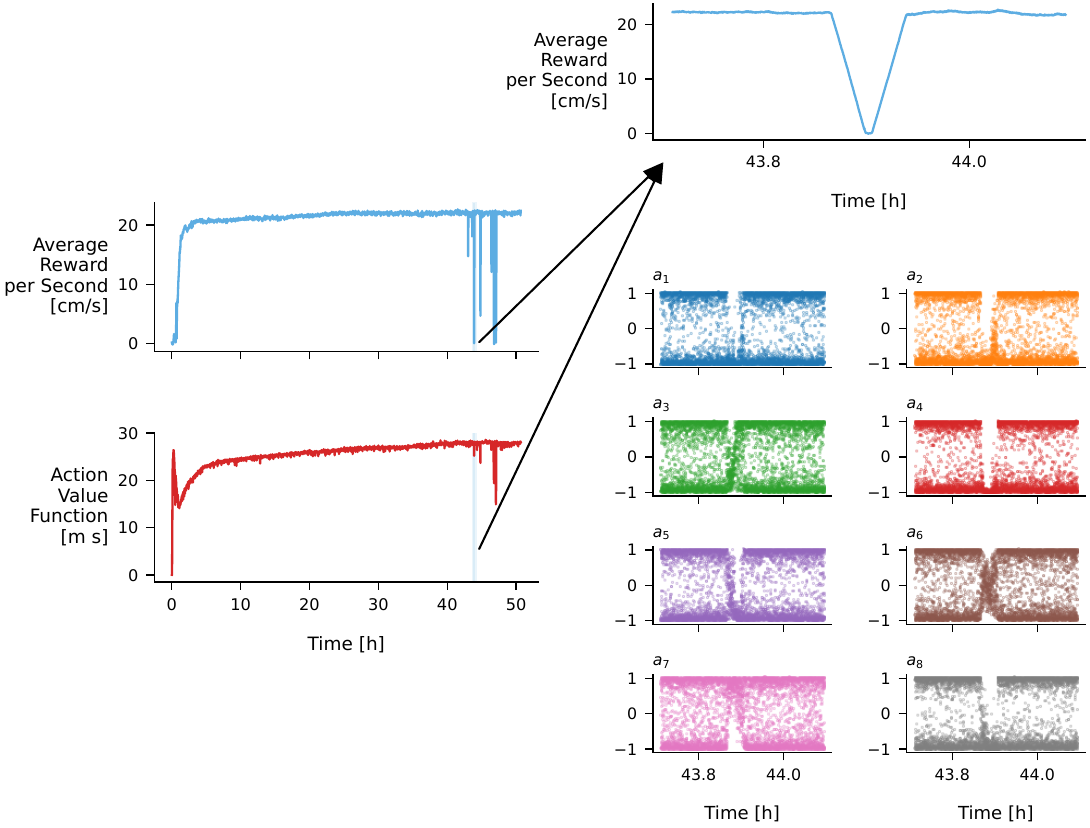}
    \caption{\textbf{Reward drop.} Performance in simulation during learning of the back-and-forth task using SAC (seed 14). \textbf{Left:} Average reward per second and the action-value function evaluated at the current state-action pair throughout learning. After approximately 40 hours of continual learning, the agent experiences a noticeable drop in performance. \textbf{Right:} A zoomed-in view of the period surrounding this performance degradation, together with the corresponding actions taken by the agent. Note that for this run, no LayerNorm was used.}
    \label{fig:figure_reward_drop}
\end{figure}

\revision{
We observed occasional runs in which the performance in simulation drops to nearly zero.
Similar behavior has also been observed on the physical robot, as illustrated in \href{https://www.youtube.com/watch?v=tHW0VfXyEPQ}{Movie 4} for SARSA($\lambda$). In both cases, the agent eventually recovers and resumes learning, as shown in \href{https://youtu.be/NCSODJeMJPw}{Movie 5} for the simulation.
The underlying cause of these temporary performance drops remains unclear and warrants further investigation. This phenomenon is illustrated in~\Cref{fig:figure_reward_drop}.
}

\cleardoublepage
\section{Step Timing}\label{sec:timing}

\begin{figure}[H]
    \centering
    \includegraphics[
        page=1,
        width=\textwidth
    ]{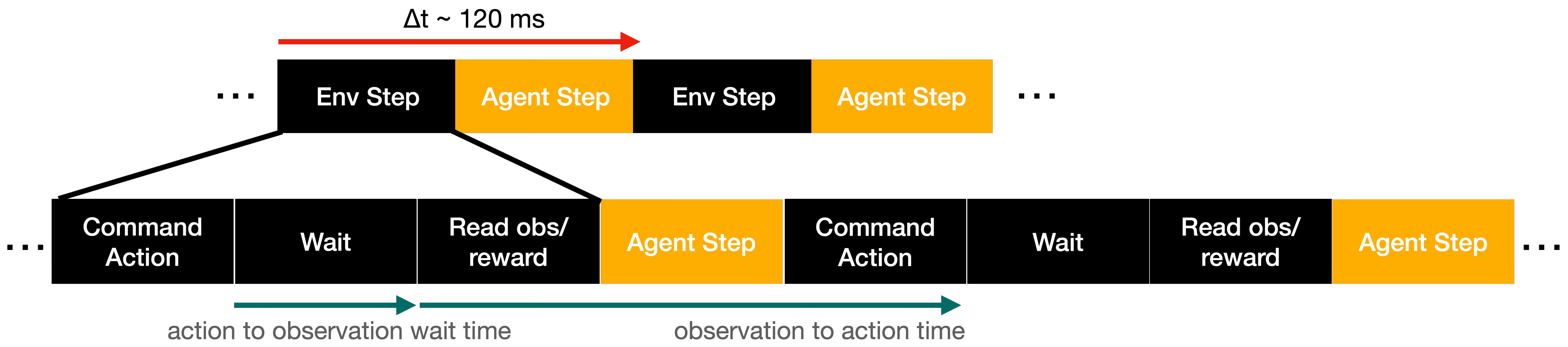}
\caption{\textbf{Step timing.} This diagram highlights the breakdown of time in an agent-environment step.}
    \label{fig:timing_env}
\end{figure}

In~\Cref{fig:timing_env}, we present a timing diagram of the interaction loop of the environment and the learning agent. 
In order to maintain a fixed step frequency, the wait time between issuing motor commands and reading observations is adjusted at every cycle.
This has the advantage of maximizing the elapsed time between applying the action and observing its result, approximating a simulated environment.
Simulated environments typically apply an action, step forward the simulation time, and then read the observation. This idealized setting compresses the observation to action time to zero, since the simulation time does not advance.

However the drawback of this scheme is that the perceived  time from action to observations is not fixed. Different strategies, like fixing action to observation wait time to a lower bound and moving the variable wait time into the observation to action part of the cycle may be explored in future work.

%% file: appendix/appendix_motion_primitives.tex
\cleardoublepage
\section{Motion Primitives for SARSA($\lambda$)}\label{sec:motion_primitives_sarsa}

\begin{figure}[H]
    \centering
    \includegraphics[width=\linewidth]{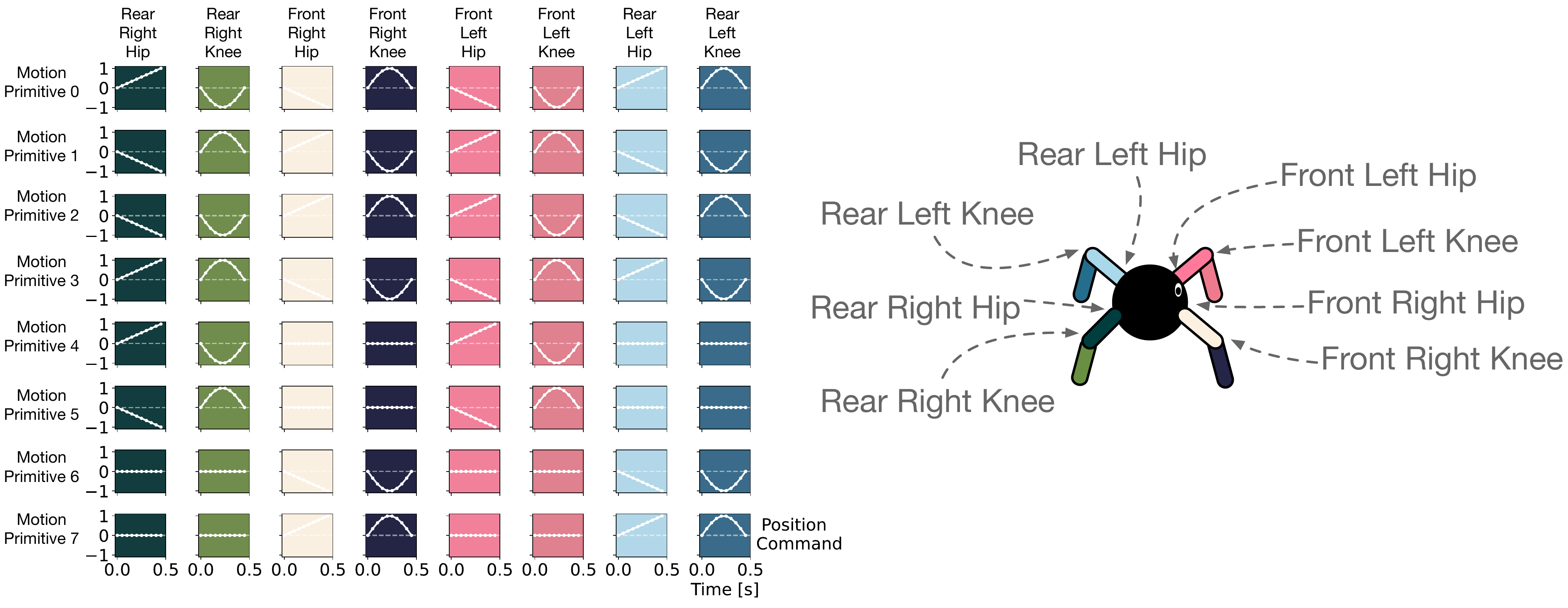}
    \caption{\textbf{Motion primitives used for the SARSA($\lambda$) \revision{algorithm}.} 
    Each color \revision{in the plot and on the robot} is associated with a leg part and position (hip or knee, left or right). 
    We chose to design 8 motion primitives in total.}
    \label{fig:motion_primitives_sarsa}
\end{figure}

%% file: appendix/appendix_list_movies.tex
\section{List of Movies Mentioned in the Submission}

\begin{itemize}
    \item \href{https://www.youtube.com/watch?v=9aowd2JxAnE}{Movie 1}. Assembly of the Open Ant.
    \item \href{https://www.youtube.com/watch?v=QJuqB7gb4y0&t=322s}{Movie 2}. Learning to walk using SAC.
    \item \href{https://www.youtube.com/watch?v=9VMjJRwFiRw}{Movie 3}. Simulation results of the robot exploiting the simulator for faster walking.
    \item \href{https://www.youtube.com/watch?v=tHW0VfXyEPQ}{Movie 4}. Performance drop (robot getting stuck) while learning on hardware with SARSA.
     \item \href{https://youtu.be/NCSODJeMJPw}{Movie 5}. Performance drop (robot getting stuck) while learning in simulation with SAC.
    \item \href{https://www.youtube.com/watch?v=w6y4_IwT_G0}{Movie 6}. Openmind Research Institute Winter School.
\end{itemize}

%% file: main.bib
@book{sutton2018reinforcement,
  title={Reinforcement Learning: An Introduction},
  author={Sutton, Richard S. and Barto, Andrew G.},
  volume={1},
  edition = {Second},
  year={2018},
  publisher={MIT Press}
}

@article{tesauro1995temporal,
  title={Temporal difference learning and {TD-G}ammon},
  author={Tesauro, Gerald},
  journal={Communications of the ACM},
  volume={38},
  number={3},
  pages={58--68},
  year={1995}
}

@article{degrave2022magnetic,
  title={Magnetic control of tokamak plasmas through deep reinforcement learning},
  author={Degrave, Jonas and others},
  journal={Nature},
  volume={602},
  number={7897},
  pages={414--419},
  year={2022},
  publisher={Nature Publishing Group UK London}
}

@article{bellemare2013arcade,
  title={The {Arcade} {Learning} {Environment}: An evaluation platform for general agents},
  author={Bellemare, Marc G. and Naddaf, Yavar and Veness, Joel and Bowling, Michael},
  journal={Journal of artificial intelligence research},
  volume={47},
  pages={253--279},
  year={2013}
}

@article{silver2016mastering,
  title={Mastering the game of {Go} with deep neural networks and tree search},
  author={Silver, David and others},
  journal={Nature},
  volume={529},
  number={7587},
  pages={484--489},
  year={2016},
  publisher={Nature Publishing Group UK London}
}

@misc{towers2025gymnasiumstandardinterfacereinforcement,
      title={Gymnasium: A Standard Interface for Reinforcement Learning Environments}, 
      author={Mark Towers and others},
      year={2025},
      eprint={2407.17032},
      archivePrefix={arXiv},
      primaryClass={cs.LG},
      url={https://arxiv.org/abs/2407.17032}, 
}

@article{bellemare2020autonomous,
  title={Autonomous navigation of stratospheric balloons using reinforcement learning},
  author={Bellemare, Marc G. and others},
  journal={Nature},
  volume={588},
  number={7836},
  pages={77--82},
  year={2020},
  publisher={Nature Publishing Group UK London}
}

@article{barto1983neuronlike,
  author={Barto, Andrew G. and Sutton, Richard S. and Anderson, Charles W.},
  journal={IEEE Transactions on Systems, Man, and Cybernetics}, 
  title={Neuronlike adaptive elements that can solve difficult learning control problems}, 
  year={1983},
  volume={SMC-13},
  number={5},
  pages={834-846},
  doi={10.1109/TSMC.1983.6313077}}

@article{montague1996framework,
  title={A framework for mesencephalic dopamine systems based on predictive Hebbian learning},
  author={Montague, P. Read and Dayan, Peter and Sejnowski, Terrence J.},
  journal={Journal of Neuroscience},
  volume={16},
  number={5},
  pages={1936--1947},
  year={1996},
  publisher={Society for Neuroscience}
}

@article{schultz1997neural,
  title={A neural substrate of prediction and reward},
  author={Schultz, Wolfram and Dayan, Peter and Montague, P. Read},
  journal={Science},
  volume={275},
  number={5306},
  pages={1593--1599},
  year={1997},
  publisher={American Association for the Advancement of Science}
}

@inproceedings{tedrake2004stochastic,
  author={Tedrake, Russ and Zhang, Teresa Weirui and Seung, H. Sebastian},
  booktitle={International Conference on Intelligent Robots and Systems (IROS)}, 
  title={Stochastic policy gradient reinforcement learning on a simple 3{D} biped}, 
  year={2004},
  volume={3},
  number={},
  pages={2849-2854 vol.3},
  doi={10.1109/IROS.2004.1389841}}

@inproceedings{wu2023daydreamer,
  title={Daydreamer: World models for physical robot learning},
  author={Wu, Philipp and Escontrela, Alejandro and Hafner, Danijar and Abbeel, Pieter and Goldberg, Ken},
  booktitle={Conference on Robot Learning},
  pages={2226--2240},
  year={2023},
}

@article{wurman2022outracing,
  title={Outracing champion {G}ran {T}urismo drivers with deep reinforcement learning},
  author={Wurman, Peter R. and others},
  journal={Nature},
  volume={602},
  number={7896},
  pages={223--228},
  year={2022},
  publisher={Nature Publishing Group UK London}
}

@misc{brushdc,
  title        = {Brush-Type {DC} Motors Handbook},
author={General Dynamics},
  howpublished = {PDF (online)},
  institution  = {General Dynamics – Ordnance \& Tactical Systems},
  year         = {2020},
  url          = {https://www.gd-ots.com/wp-content/uploads/2020/07/Brush-Type-DC-Motors-Handbook_NAV.pdf},
  note         = {Accessed: 2025-11-26}
}

@article{huang2022cleanrl,
  author  = {Shengyi Huang and others},
  title   = {{CleanRL:} High-quality Single-file Implementations of Deep Reinforcement Learning Algorithms},
  journal = {Journal of Machine Learning Research},
  year    = {2022},
  volume  = {23},
  number  = {274},
  pages   = {1--18},
  url     = {http://jmlr.org/papers/v23/21-1342.html}
}

@misc{SpinningUp2018,
    author = {Achiam, Joshua},
    title = {{Spinning Up in Deep Reinforcement Learning}},
    year = {2018}
}

@article{sac,
  author       = {Tuomas Haarnoja and
                  Aurick Zhou and
                  Pieter Abbeel and
                  Sergey Levine},
  title        = {{Soft Actor-Critic}: {O}ff-policy Maximum Entropy Deep Reinforcement Learning
                  with a Stochastic Actor},
  journal      = {CoRR},
  volume       = {abs/1801.01290},
  year         = {2018},
  url          = {http://arxiv.org/abs/1801.01290},
  eprinttype    = {arXiv},
  eprint       = {1801.01290},
  bibsource    = {dblp computer science bibliography, https://dblp.org}
}

@book{slotine1991applied,
	address = {Englewood Cliffs, N.J},
	title = {Applied nonlinear control},
	isbn = {978-0-13-040890-7},
	publisher = {Prentice Hall},
	author = {Slotine, Jean-Jacques E. and Li, Weiping},
	year = {1991},
}

@article{O_Connell_2022,
	doi = {10.1126/scirobotics.abm6597},
  
	url = {https://doi.org/10.1126%2Fscirobotics.abm6597},
  
	year = 2022,
	month = may,
  
	publisher = {Amer. Assoc. for the Advancement of Science ({AAAS})},
  
	volume = {7},
  
	number = {66},
  
	author = {Michael O'Connell and others},
  
	title = {{Neural-Fly} enables rapid learning for agile flight in strong winds},
  
	journal = {Science Robotics}
}

@misc{smith2022walkparklearningwalk,
      title={A Walk in the Park: Learning to Walk in 20 Minutes With Model-Free Reinforcement Learning}, 
      author={Laura Smith and Ilya Kostrikov and Sergey Levine},
      year={2022},
      eprint={2208.07860},
      archivePrefix={arXiv},
      primaryClass={cs.RO},
      url={https://arxiv.org/abs/2208.07860}, 
}

@misc{DBLP:journals/corr/abs-2011-03085,
  author       = {Rinu Boney and
                  Jussi Sainio and
                  Mikko Kaivola and
                  Arno Solin and
                  Juho Kannala},
  title        = {{RealAnt:} An Open-Source Low-Cost Quadruped for Research in Real-World
                  Reinforcement Learning},
  volume       = {abs/2011.03085},
  year         = {2020},
  url          = {https://arxiv.org/abs/2011.03085},
  eprinttype    = {arXiv},
  eprint       = {2011.03085},
  bibsource    = {dblp computer science bibliography, https://dblp.org}
}

@article{berrueta2024maximum,
  title={Maximum diffusion reinforcement learning},
  author={Berrueta, Thomas A. and Pinosky, Alex and Murphey, Timothy D.},
  journal={Nature Machine Intelligence},
  volume={6},
  pages={558--567},
  year={2024},
  publisher={Nature Publishing Group}
}

@INPROCEEDINGS{robot_dog_peter_stone,
  author={Kohl, Nate and Stone, Peter},
  booktitle={IEEE International Conference on Robotics and Automation, 2004. Proceedings. ICRA '04. 2004}, 
  title={Policy gradient reinforcement learning for fast quadrupedal locomotion}, 
  year={2004},
  volume={3},
  number={},
  pages={2619-2624 Vol.3},
  doi={10.1109/ROBOT.2004.1307456}
}

@article{Schulman2015HighDimensionalCC,
  title={High-Dimensional Continuous Control Using {G}eneralized {A}dvantage {E}stimation},
  author={John Schulman and Philipp Moritz and Sergey Levine and Michael I. Jordan and Pieter Abbeel},
  journal={CoRR},
  year={2015},
  volume={abs/1506.02438},
  url={https://api.semanticscholar.org/CorpusID:3075448}
}

@INPROCEEDINGS{7989376,
  author={Preiss, James A. and Honig, Wolfgang and Sukhatme, Gaurav S. and Ayanian, Nora},
  booktitle={IEEE International Conference on Robotics and Automation}, 
  title={Crazyswarm: A large nano-quadcopter swarm}, 
  year={2017},
  volume={},
  number={},
  pages={3299-3304},
  doi={10.1109/ICRA.2017.7989376}}

@misc{preiss2025fastnonepisodicadaptivetuning,
      title={Fast Non-Episodic Adaptive Tuning of Robot Controllers with Online Policy Optimization}, 
      author={James A. Preiss and Fengze Xie and Yiheng Lin and Adam Wierman and Yisong Yue},
      year={2025},
      eprint={2507.10914},
      archivePrefix={arXiv},
      primaryClass={cs.RO},
      url={https://arxiv.org/abs/2507.10914}, 
}

@misc{ba2016layernormalization,
      title={Layer Normalization}, 
      author={Jimmy Lei Ba and Jamie Ryan Kiros and Geoffrey E. Hinton},
      year={2016},
      eprint={1607.06450},
      archivePrefix={arXiv},
      primaryClass={stat.ML},
      url={https://arxiv.org/abs/1607.06450}, 
}

@inproceedings{optuna, author = {Akiba, Takuya and Sano, Shotaro and Yanase, Toshihiko and Ohta, Takeru and Koyama, Masanori}, title = {Optuna: A Next-generation Hyperparameter Optimization Framework}, year = {2019}, doi = {10.1145/3292500.3330701},  booktitle = {Proceedings of the 25th ACM SIGKDD International Conference on Knowledge Discovery \& Data Mining}, pages = {2623–2631}, numpages = {9}}

@INPROCEEDINGS{olson-april-11,
  author={Olson, Edwin},
  booktitle={2011 IEEE International Conference on Robotics and Automation}, 
  title={April{T}ag: A robust and flexible visual fiducial system}, 
  year={2011},
  volume={},
  number={},
  pages={3400-3407},
  doi={10.1109/ICRA.2011.5979561}}

@misc{mahmood2018settingreinforcementlearningtask,
      title={Setting up a Reinforcement Learning Task with a Real-World Robot}, 
      author={A. Rupam Mahmood and Dmytro Korenkevych and Brent J. Komer and James Bergstra},
      year={2018},
      eprint={1803.07067},
      archivePrefix={arXiv},
      primaryClass={cs.LG},
      url={https://arxiv.org/abs/1803.07067}, 
}

@article{DBLP:journals/corr/abs-1809-07731,
  author       = {A. Rupam Mahmood and
                  Dmytro Korenkevych and
                  Gautham Vasan and
                  William Ma and
                  James Bergstra},
  title        = {Benchmarking Reinforcement Learning Algorithms on Real-World Robots},
  journal      = {CoRR},
  volume       = {abs/1809.07731},
  year         = {2018},
  url          = {http://arxiv.org/abs/1809.07731},
  eprinttype    = {arXiv},
  eprint       = {1809.07731},
  bibsource    = {dblp computer science bibliography, https://dblp.org}
}

@misc{watkins_thesis,
author = {Watkins, Christopher},
year = {1989},
pages = {},
title = {Learning From Delayed Rewards},
url = {https://www.cs.rhul.ac.uk/~chrisw/new_thesis.pdf}
}

@misc{elsayed2024streamingdeepreinforcementlearning,
      title={Streaming Deep Reinforcement Learning Finally Works}, 
      author={Mohamed Elsayed and Gautham Vasan and A. Rupam Mahmood},
      year={2024},
      eprint={2410.14606},
      archivePrefix={arXiv},
      primaryClass={cs.LG},
      url={https://arxiv.org/abs/2410.14606}, 
}

@misc{sutton2025oak,
  author       = {Richard S. Sutton},
  title        = {The {OaK} Architecture: A Vision of {SuperIntelligence} from Experience},
  year         = {2025},
  howpublished = {Keynote talk presented at the Reinforcement Learning Conference (RLC), 2025},
  url = {https://rl-conference.cc/2025/keynotes.html},
}

@misc{kaelbling2025rational,
  author       = {Leslie Kaelbling},
  title        = {R{L}: {R}ational {L}earning},
  year         = {2025},
  howpublished = {Keynote talk presented at the Reinforcement Learning Conference (RLC), 2025},
  url = {https://rl-conference.cc/2025/keynotes.html},
}

@article{DBLP:journals/corr/abs-1812-06298,
  author       = {Tom Silver and
                  Kelsey R. Allen and
                  Josh Tenenbaum and
                  Leslie Pack Kaelbling},
  title        = {Residual Policy Learning},
  journal      = {CoRR},
  volume       = {abs/1812.06298},
  year         = {2018},
  url          = {http://arxiv.org/abs/1812.06298},
  eprinttype    = {arXiv},
  eprint       = {1812.06298},
  bibsource    = {dblp computer science bibliography, https://dblp.org}
}

@article{tilecoding_james_albus,
  author = {James Albus},
  title = {New Approach to Manipulator Control: {T}he {C}erebellar {M}odel {A}rticulation {C}ontroller {(CMAC)}},
  year = {1975},
  month = {1975-09-30},
  journal = {Transactions of the ASME Journal of Dynamic Systems},
  url = {https://tsapps.nist.gov/publication/get_pdf.cfm?pub_id=820151},
  language = {en},
}

@inproceedings{NIPS1995_8f1d4362,
 author = {Sutton, Richard S},
 booktitle = {Advances in Neural Information Processing Systems},
 pages = {},
 publisher = {MIT Press},
 title = {Generalization in Reinforcement Learning: Successful Examples Using Sparse Coarse Coding},
 url = {https://proceedings.neurips.cc/paper_files/paper/1995/file/8f1d43620bc6bb580df6e80b0dc05c48-Paper.pdf},
 volume = {8},
 year = {1995}
}

@article{domain_randomization,
  author       = {Joshua Tobin and
                  Rachel Fong and
                  Alex Ray and
                  Jonas Schneider and
                  Wojciech Zaremba and
                  Pieter Abbeel},
  title        = {Domain Randomization for Transferring Deep Neural Networks from Simulation
                  to the Real World},
  journal      = {CoRR},
  volume       = {abs/1703.06907},
  year         = {2017},
  url          = {http://arxiv.org/abs/1703.06907},
  eprinttype    = {arXiv},
  eprint       = {1703.06907},
  bibsource    = {dblp computer science bibliography, https://dblp.org}
}

@misc{deasis2024idiosyncrasytimediscretizationreinforcementlearning,
      title={An Idiosyncrasy of Time-discretization in Reinforcement Learning}, 
      author={De Asis, Kris and Sutton, Richard S. },
      year={2024},
      eprint={2406.14951},
      archivePrefix={arXiv},
      primaryClass={cs.LG},
      url={https://arxiv.org/abs/2406.14951}, 
}

@article{10.1109/TRO.2024.3475212,
author = {Lupu, Elena Sorina and Xie, Fengze and Preiss, James Alan and Alindogan, Jedidiah and Anderson, Matthew and Chung, Soon-Jo},
title = {{MAGIC$^\mathrm{{VFM}}$}-{M}eta-Learning Adaptation for Ground Interaction Control With Visual Foundation Models},
year = {2025},
issue_date = {2025},
publisher = {IEEE Press},
volume = {41},
issn = {1552-3098},
url = {https://doi.org/10.1109/TRO.2024.3475212},
doi = {10.1109/TRO.2024.3475212},
journal = {IEEE Transactions on Robotics},
month = jan,
pages = {180–199},
numpages = {20}
}

@inproceedings{Shi2020Adaptive,
  author    = {Xichen Shi and Patrick Spieler and Ellande Tang and Elena-Sorina Lupu and Phillip Tokumaru and Soon-Jo Chung},
  title     = {Adaptive Nonlinear Control of Fixed-Wing {VTOL} with Airflow Vector Sensing},
  booktitle = {2020 IEEE International Conference on Robotics and Automation (ICRA)},
  year      = {2020},
  pages      = {5321--5327},
  doi        = {10.1109/ICRA40945.2020.9197344},
}

@inproceedings{10.1109/IROS.2018.8593894,
author = {Rupam Mahmood, A. and Korenkevych, Dmytro and Komer, Brent J. and Bergstra, James},
title = {Setting up a Reinforcement Learning Task with a Real-World Robot},
year = {2018},
publisher = {IEEE Press},
url = {https://doi.org/10.1109/IROS.2018.8593894},
doi = {10.1109/IROS.2018.8593894},
booktitle = {2018 IEEE/RSJ International Conference on Intelligent Robots and Systems (IROS)},
pages = {4635–4640},
numpages = {6},
location = {Madrid, Spain}
}

@inproceedings{yuan2022asynchronous,
  author    = {Yuan, Yufeng and Mahmood, A. Rupam},
  title     = {Asynchronous Reinforcement Learning for Real-Time Control of Physical Robots},
  booktitle = {IEEE International Conference on Robotics and Automation (ICRA)},
  year      = {2022}
}

@InProceedings{pmlr-v330-farrahi26a,
  title = 	 {Learning Without Time-Based Embodiment Resets in Soft-Actor Critic},
  author =       {Farrahi, Homayoon and Mahmood, A. Rupam},
  booktitle = 	 {Proceedings of The 4th Conference on Lifelong Learning Agents},
  pages = 	 {112--135},
  year = 	 {2026},
  volume = 	 {330},
  series = 	 {Proceedings of Machine Learning Research},
  month = 	 {11--14 Aug},
  publisher =    {PMLR},
  url = 	 {https://proceedings.mlr.press/v330/farrahi26a.html}
}

@article{doi:10.1126/scirobotics.abk2822,
author = {Takahiro Miki  and Joonho Lee  and Jemin Hwangbo  and Lorenz Wellhausen  and Vladlen Koltun  and Marco Hutter },
title = {Learning robust perceptive locomotion for quadrupedal robots in the wild},
journal = {Science Robotics},
volume = {7},
number = {62},
pages = {eabk2822},
year = {2022},
doi = {10.1126/scirobotics.abk2822},

URL = {https://www.science.org/doi/abs/10.1126/scirobotics.abk2822},
eprint = {https://www.science.org/doi/pdf/10.1126/scirobotics.abk2822},
}

@article{
doi:10.1126/science.adw1291,
author = {Jonas Buchli  and others},
title = {Improving cosmological reach of a gravitational wave observatory using Deep Loop Shaping},
journal = {Science},
volume = {389},
number = {6764},
pages = {1012-1015},
year = {2025},
doi = {10.1126/science.adw1291},
URL = {https://www.science.org/doi/abs/10.1126/science.adw1291},
eprint = {https://www.science.org/doi/pdf/10.1126/science.adw1291}}

@INPROCEEDINGS{287219,
  author={Benbrahim, H. and Doleac, J.S. and Franklin, J.A. and Selfridge, O.G.},
  booktitle={[Proceedings 1992] IJCNN International Joint Conference on Neural Networks}, 
  title={Real-time learning: a ball on a beam}, 
  year={1992},
  volume={1},
  number={},
  pages={98-103 vol.1},
  doi={10.1109/IJCNN.1992.287219}}
